\documentclass[a4paper]{cas-sc}
\usepackage{graphicx}   
\usepackage{subcaption} 

\usepackage[numbers]{natbib}

\def\tsc#1{\csdef{#1}{\textsc{\lowercase{#1}}\xspace}}
\tsc{WGM}
\tsc{QE}

\begin{document}
\let\WriteBookmarks\relax
\def\floatpagepagefraction{1}
\def\textpagefraction{.001}
\let\printorcid\relax 

\shorttitle{3D Trajectory Reconstruction of Moving Points Based on a Monocular Camera}   

\shortauthors{Huang et al.}

\title[mode = title]{3D Trajectory Reconstruction of Moving Points Based on a Monocular Camera}  

\author[1,2]{Huayu Huang}

\author[1,2]{Banglei Guan}

\author[1,2]{Yang Shang}

\author[1,2]{Qifeng Yu}

\address[1]{College of Aerospace Science and Engineering, National University of Defense Technology, Changsha, China}
\address[2]{Hunan Provincial Key Laboratory of Image Measurement and Vision Navigation, National University of Defense Technology, Changsha, China}

\begin{abstract}
The motion measurement of point targets constitutes a fundamental problem in photogrammetry, with extensive applications across various engineering domains. Reconstructing a point's 3D motion just from the images captured by only a monocular camera is unfeasible without prior assumptions. Under limited observation conditions such as insufficient observations, long distance, and high observation error of platform, the least squares estimation faces the issue of ill-conditioning. This paper presents an algorithm for reconstructing 3D trajectories of moving points using a monocular camera. The motion of the points is represented through temporal polynomials. Ridge estimation is introduced to mitigate the issues of ill-conditioning caused by limited observation conditions. Then, an automatic algorithm for determining the order of the temporal polynomials is proposed. Furthermore, the definition of \textit{reconstructability} for temporal polynomials is proposed to describe the reconstruction accuracy quantitatively. The simulated and real-world experimental results demonstrate the feasibility, accuracy, and efficiency of the proposed method.
\end{abstract} 



\maketitle

\section{Introduction}\label{sec1}

UAVs stand out because of their many advantages, such as economy, remote control, and no casualties. It is widely used in smart cities \cite{Mohamed2020,Lee2024}, damage detection \cite{Feng2024,Liang2023}, precision agriculture \cite{Xiao2023,Subeesh2024}, and other fields \cite{Guan2021,Erol2024,Wei2024}. Localizing terrestrial or maritime targets is crucial to numerous unmanned aerial vehicle (UAV) applications \cite{Guan2018}. Almost all of these applications require rapid high-precision localization of the target of interest. In photogrammetry, binocular stereo is usually used to reconstruct a 3D point \cite{Guan2022,Bian2024}. The conventional triangulation approach is geometrically sound and well-defined. Two sight-rays connecting the optical centers of the two cameras along with the baseline form a triangle, and the intersection of the two sight-rays is the position of the target point. This process is called triangulation. The triangulation technology of binocular vision has been systematically developed \cite{Longuet1981,Hartley2004,Lee2019}. 

\begin{figure}[htbp]  
\centering
\subfloat[]{\includegraphics[width=1.8in]{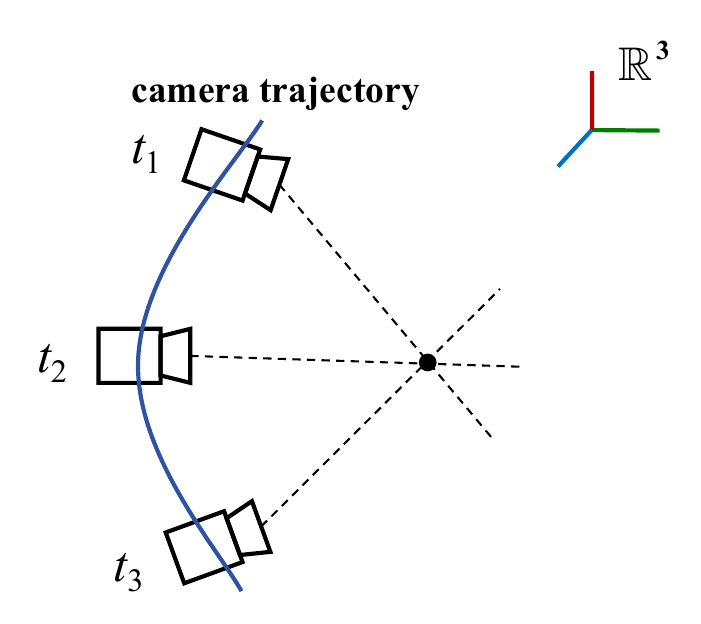}%
\label{fig1a}}
\hfil
\subfloat[]{\includegraphics[width=1.8in]{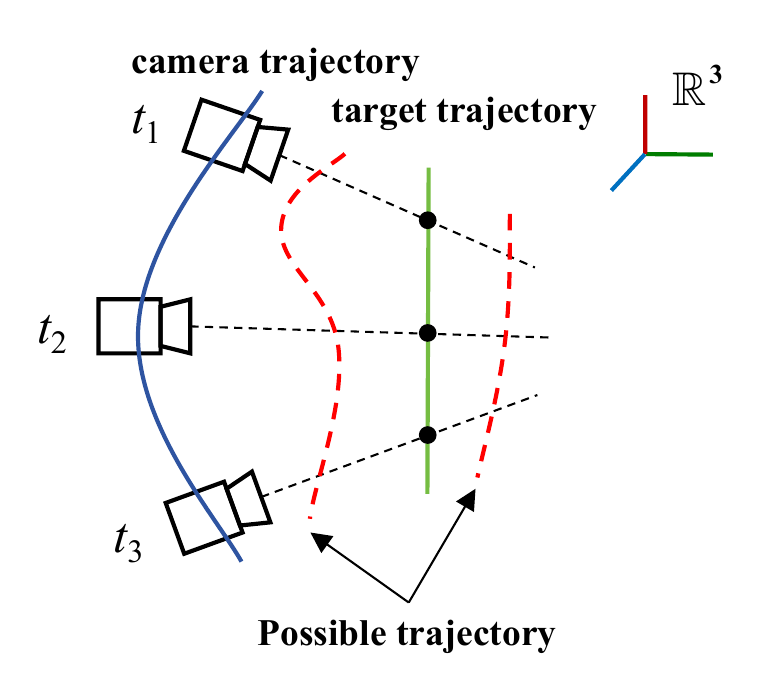}%
\label{fig1b}}
\caption{3D reconstruction of a point based on a monocular camera. (a) A static point. (b) A moving point.}
\label{fig1}
\end{figure}

Recently, monocular camera based target localization techniques have received considerable attention due to their advantages of low cost, light weight, and power efficiency \cite{Zhang2020,Wang2022,Yu2024_1}. Due to the limitations of UAV platforms, often only a monocular camera can be installed. However, it is still possible to realize the triangulation of static points by moving the camera, as shown in Figure \ref{fig1a}. We can also estimate the depth through deep learning, which is currently the most popular method \cite{Xu2021,Wan2024,Liu2024}. Unfortunately, the geometric constraint of triangulation becomes inapplicable when the point moves between image acquisitions, as shown in Figure \ref{fig1b}. This situation is very common, because most artificial vision systems are monocular, and most real scenes contain moving targets \cite{Tan2013,Saputra2018}. In this case, without reasonable assumptions about the target's motion or assistance of other sensors \cite{Yin2022,Zhang2023,Lan2024}, it is impossible to reconstruct the 3D trajectory of the target. Compared with relying on the assistance of other sensors, the method based on motion assumptions has the advantages of being lighter, more economical, and more versatile in applicability. Based on the background in this paper, the trajectories of point targets are reconstructed using motion assumptions.

There are three types of methods based on motion assumptions of the moving points. The first type is called trajectory triangulation, which assumes the shape of the target trajectory as an additional constraint. The principal work of trajectory triangulation is by Avidan and Shashua \cite{Avidan2000}. They assume that the target point moves along a straight line \cite{Avidan1999} or a conic \cite{Shashua1999}. The limitation of this method is that the shape of the target trajectory must be known in advance, and it must be a line or a conic section. However, their method has inspired numerous trajectory recovery algorithms based on shape constraints. Based on this work, Shashua et al. \cite{Shashua2000,Wexler2000} introduce the dual Htensor, a generalization of the well-known homography tensor. It allows for solving the trajectory of a point moving along a straight line with only three views and does not require prior information on camera positions. However, this method can only solve points moving along a straight line. Wolf and Shashua \cite{Wolf2002} then address the geometry of multiple views of dynamic scenes by lifting the problem to a static scene embedded in a higher dimensional space of $\mathbb{P}^N$. Their work explores six applications utilizing different $N$ values between 3 and 6. Kaminski and Teicher \cite{Kaminski2004} propose a general framework for trajectory triangulation by extending these ideas to a family of hyper-surfaces in the projective space $\mathbb{P}^5$. This approach enables the reconstruction of diverse trajectories expressible as polynomials. However, this framework is sensitive to noise, and the number of equations grows exponentially. The disadvantage of this type of method is that acquiring an initial value is not easy, and an initial value without enough precision will lead to failure of subsequent iteration. Generally, these methods can only get good results with quite good initial values.

The second type of method is to represent 3D trajectories using linear combinations of trajectory basis vectors such that the recovery of 3D points can be estimated robustly using least squares. Akhter et al. \cite{Akhter2008,Akhter2011} propose discrete cosine transform (DCT) as a trajectory basis to estimate non-rigid structures. Any motion can be expressed by this trajectory representation, which operates independently of specific objects and requires no prior knowledge. However, it cannot handle missing information. Park et al. \cite{Park2010} propose a method based on the DCT trajectory basis vectors to recover 3D trajectories of moving points from a series of 2D projections. This method can reconstruct arbitrary complex motions such as human motion and handle missing information. They also propose an index called \textit{reconstructability} to describe the reconstruction accuracy of the system quantitatively. Subsequently, Park et al. \cite{Park2015} further propose an automated method for selecting the number of trajectory basis vectors. Based on the DCT trajectory basis vectors, Zhu et al. \cite{Zhu2011} propose to introduce the L1 regularization term. They also select some keyframes to calculate the position of points independently as prior knowledge \cite{Wei2009}. Their approach refines 3D reconstruction precision while diminishing the correlation between camera and point trajectories. However, the reconstruction method used in keyframes requires prior information on human joints, which greatly limits the application domain. This type of method based on DCT trajectory basis vectors is mainly aimed at non-rigid motions. Its disadvantage is a lack of clear physical meaning so it needs to solve an excessive number of parameters and performs poor accuracy for simple motions. Moreover, when selecting the number of trajectory basis vectors, there are too many iterations, resulting in a long computing time. This is unnecessary for solving simple motions of vehicle or ship targets.

The third type of method is called trajectory intersection, which represents the motion of the target points as temporal polynomials. Zhang and Yu et al. \cite{Zhang2006,Yu2009} propose the principal work of the trajectory intersection. It is based on the collinearity equation and the motion of the point is represented as temporal polynomials. Since temporal polynomials have significant physical meaning, this method is particularly suitable for targets undergoing simple mechanical motion within a certain period of time. Therefore, The method performs excellently in estimating the trajectories of vehicles and ships. Li et al. \cite{Li2014} introduce a kernel function to the trajectory intersection method and propose a motion measurement method of a point target based on the least square support vector machines (LSSVM). This method has higher accuracy under high noise levels because of the addition of the kernel function. The incremental and decremental technique is used to improve computational efficiency. Li et al. \cite{Li2015} also extended the trajectory intersection method to multi-camera systems. Compared to classical two-view triangulation, this method can handle missing information and the conditions of asynchrony, no time registration, or even no time information among cameras. Zhou et al. \cite{Zhou2015} conduct a detailed analysis of the trajectory intersection method and present the necessary and sufficient conditions for the uniqueness of its solution. They also extend trajectory intersection to capture the orientation of the moving object, which PnP methods would not obtain due to a lack of features. Chen et al. \cite{Chen2019} aimed at maritime targets, utilizing a dynamic sea surface elevation model to enhance maritime target motion constraints. This approach simplifies the target's 3D spatial motion to a 2D surface motion to achieve higher accuracy. The advantage of this type of method is that it has a clear physical meaning and high calculation efficiency. However, the 3D reconstruction accuracy is low or even degenerates due to serious ill-conditioning caused by limited observation conditions. The order of the temporal polynomials controls the complexity of the reconstructed trajectory and the number of parameters to be estimated. Selecting an appropriate order of temporal polynomials is essential for achieving high-precision measurements of the target motion. In current methods, the order parameter is manually selected based on experience, which often leads to low accuracy.

This paper focuses on the 3D trajectory reconstruction of ground-moving targets, such as vehicles and ships, using UAVs. It is impossible to measure the 3D position of a moving point solely based on the images captured by a monocular camera without making reasonable assumptions about the point’s motion. This paper proposes a 3D trajectory reconstruction method for moving points using a monocular camera. First, temporal polynomials are employed to represent the point’s motion, and ridge estimation is introduced into the least squares estimation system to mitigate the ill-conditioning. Second, an automatic algorithm for determining the order of the temporal polynomials is proposed by minimizing a geometric error objective function. Third, the \textit{reconstructability} for temporal polynomials is defined by analyzing the geometric relationships among camera motion, target point motion, and the temporal polynomial, serving as a quantitative index. Finally, the efficiency, accuracy, and robustness of the proposed method are validated through both simulated and real-world data.

The main contributions of this paper include:
\begin{enumerate}[$\bullet$]
    \item We use temporal polynomials to represent the point motion as an additional constraint. Ridge estimation is introduced to the least squares estimation system to mitigate the ill-conditioning caused by limited observation conditions, thereby improving the accuracy and robustness.
      
    \item We propose an efficient automatic selection algorithm for the order of the temporal polynomials. The proposed algorithm automatically determines the optimal order of the temporal polynomials by minimizing an objective function of geometric error.
      
    \item We define the \textit{reconstructability} for temporal polynomials by analyzing the geometric relationships among camera motion, target point motion, and the temporal polynomial. This index is then used to describe the reconstruction accuracy quantitatively.
\end{enumerate}  

The rest of the paper is structured as follows: Section \ref{sec2} proposes the 3D trajectory reconstruction method for moving points and the automatic selection algorithm for the order of the temporal polynomials. Furthermore, Section \ref{sec2} discusses the geometric relationships among camera motion, target motion, and temporal polynomials. Sections \ref{sec3} and \ref{sec4} respectively validate the proposed method through simulated and real-world experiments. Ultimately, Section \ref{sec5} offers a conclusion. 
\section{Methods}\label{sec2}

This section proposes the 3D trajectory reconstruction method of moving points, as well as the automatic selection method of the order parameter. Then a detailed geometric analysis is conducted on the reconstructability of the algorithm.

\subsection{Linear reconstruction of a 3D point trajectory}\label{sec2.1}

\begin{figure}[htbp]
\centering
\includegraphics[width=3in]{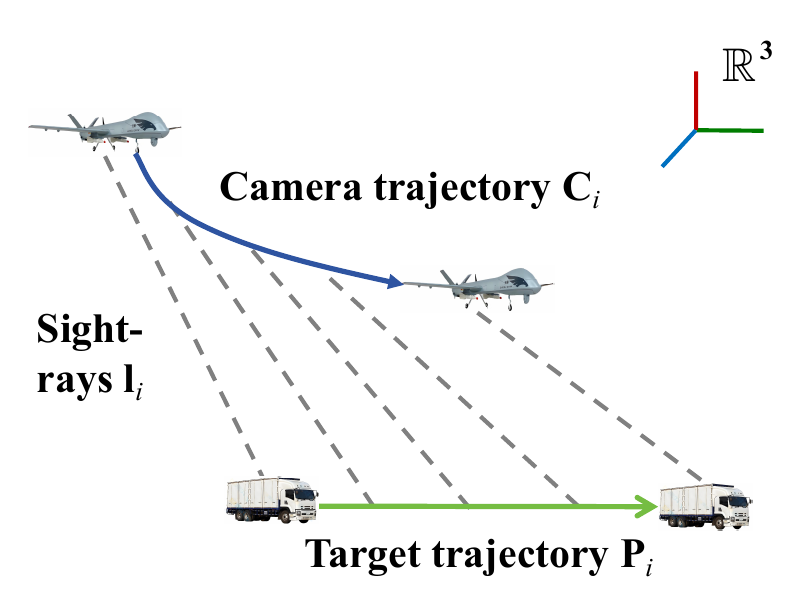}
\caption{Illustration of 3D trajectory reconstruction of a moving target based on a flight platform equipped with a monocular camera.}
\label{fig2}
\end{figure}

Figure \ref{fig2} illustrates the scenario where a flight platform, equipped with a monocular camera, is observing a moving target continually. The target can be regarded as a point target. Assuming that at the time $t_i$, the position of the camera is \(\mathbf{C}_i=\begin{bmatrix}X_{Ci}&Y_{Ci}&Z_{Ci}\end{bmatrix}^\mathrm{T}\), which is accurately known. The position of the target is \(\mathbf{P}_i=\begin{bmatrix}X_i&Y_i&Z_i\end{bmatrix}^\mathrm{T}\). Let $\mathbf{P}_{i}$ be imaged as \(\mathbf{p}_i=\begin{bmatrix}x_i&y_i\end{bmatrix}^\mathrm{T}\). The internal parameter $\mathbf{K}$ is calibrated in advance\cite{Hartley2004,Liang2024}. The rotation matrix $\mathbf{R}_{i}$ of the camera can be calculated employing the technique of structure from motion within a given scene \cite{Guan2023,Yu2024_2}. Therefore, we can calculate the direction of the observation sight-ray \(\mathbf{l}_i=\begin{bmatrix}l_{Xi}&l_{Yi}&l_{Zi}\end{bmatrix}^\mathrm{T}\),
\begin{equation}
    \mathbf{l}_i=\frac{\mathbf{R}_i^\mathrm{T}\mathbf{K}^{-1}\mathbf{p}_i}{\|\mathbf{R}_i^\mathrm{T}\mathbf{K}^{-1}\mathbf{p}_i\|}.
    \label{eq1}
\end{equation} 

\begin{figure}[htbp]
\centering
\includegraphics[width=2.2in]{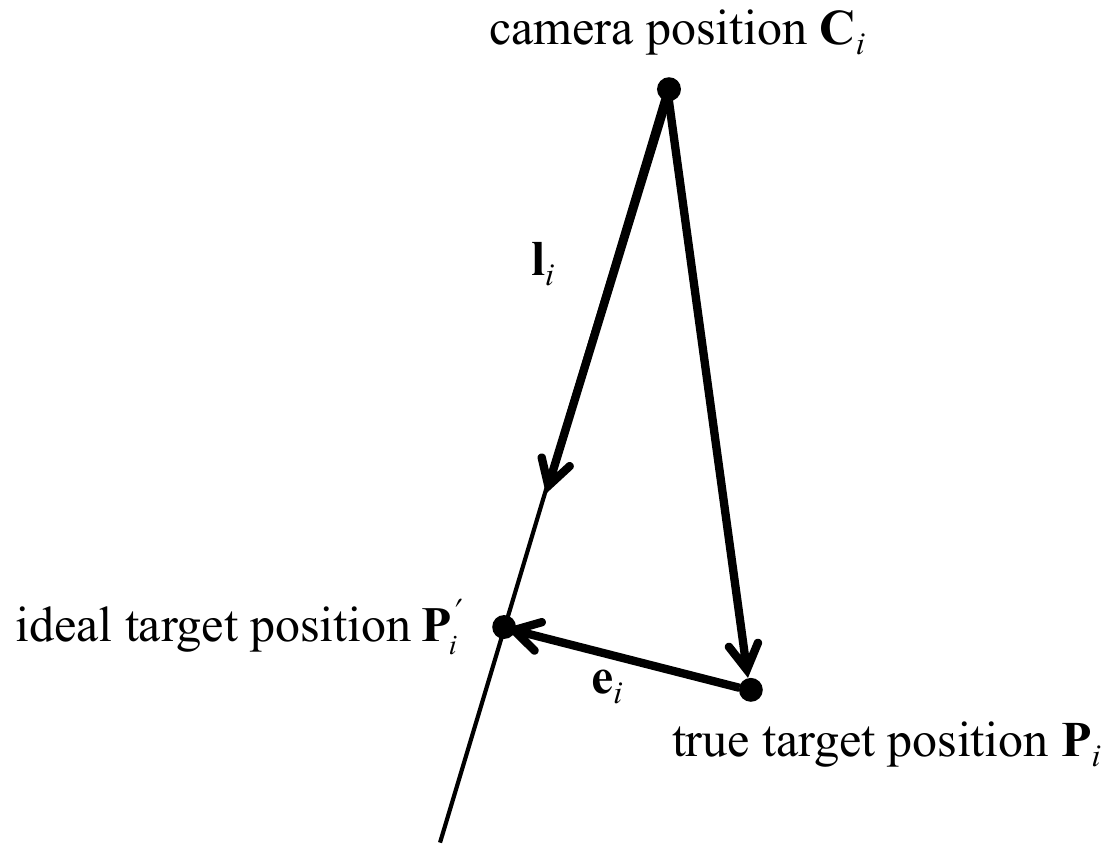}
\caption{Illustration of residual error.}
\label{fig3}
\end{figure}

As shown in Figure \ref{fig3}, the residual error $\mathbf{e}_{i}$ between the true position and the ideal position of the target can be represented as: 
\begin{equation}  
    \mathbf{e}_{i} = (\mathbf{I} - \mathbf{l}_{i}\mathbf{l}_{i}^\mathrm{T} )(\mathbf{C}_{i} - \mathbf{P}_{i}),
    \label{eq2}
\end{equation}  
where $\mathbf{I}$ represents the identity matrix. Under the criterion of minimizing the sum of squared residuals, we can establish the following equation: 
\begin{equation}
    (\mathbf{I}-\mathbf{l}_{i}\mathbf{l}_{i}^\mathrm{T})\mathbf{P}_{i}=(\mathbf{I}-\mathbf{l}_{i}\mathbf{l}_{i}^\mathrm{T})\mathbf{C}_{i},
    \label{eq3}
\end{equation} 
where the rank of this equation is 2. Therefore, by combining Eq. (\ref{eq3}) from $N$ observations, the number of unknowns is 3$N$, while the number of independent equations is $2N$. For a specific set of measurements, there exists an infinite array of solutions.

To calculate the position of the target, additional constraints must be imposed on its trajectory. We achieve this by expressing the target trajectory as polynomials of time $t_i$:
\begin{equation}
    \left(X_{i}=\sum\limits_{k=0}^{K}a_{k}t_i^{k}\quad Y_{i}=\sum\limits_{k=0}^{K}b_{k}t_i^{k}\quad Z_{i}=\sum\limits_{k=0}^{K}c_{k}t_i^{k}\right),
    \label{eq4}
\end{equation} 
where $K$ is the order of the polynomials. $a_{k}(k=0, 1,\cdots, K)$, $b_{k}(k=0, 1,\cdots, K)$, $c_{k}(k=0, 1,\cdots, K)$ are the motion parameters of the target that need to be solved. Based on Eq. (\ref{eq3}) and Eq. (\ref{eq4}), the subsequent set of equations can be formulated:
\begin{equation}
    (\mathbf{I}-\mathbf{l}_{i}\mathbf{l}_{i}^\mathrm{T})\begin{bmatrix}\sum\limits_{k=0}^{K}a_{k}t_i^{k}\\\sum\limits_{k=0}^{K}b_{k}t_i^{k}\\\sum\limits_{k=0}^{K}c_{k}t_i^{k}\end{bmatrix}=(\mathbf{I}-\mathbf{l}_{i}\mathbf{l}_{i}^\mathrm{T})\mathbf{C}_{i}.
    \label{eq5}
\end{equation}

Solving Eq. (\ref{eq5}) from $N$ observations for the motion parameters of the target,  the number of unknowns is $3(K+1)$, while the number of independent equations is $2N$. It is a linear least squares system if $2N \geq 3(K+1)$. The estimated motion parameters $a_{k}$, $b_{k}$, and $c_{k}$ can be obtained by solving the linear least equations. The reconstructed trajectory of the target is a trajectory that passes through all sight-rays and is represented by linear polynomials of time $t_i$. Then the moving trajectory of the target can be reconstructed by Eq. (\ref{eq4}).

\subsection{Trajectory reconstruction under ridge estimation}\label{sec2.2}

\begin{figure}[htbp]
\centering
\includegraphics[width=5in]{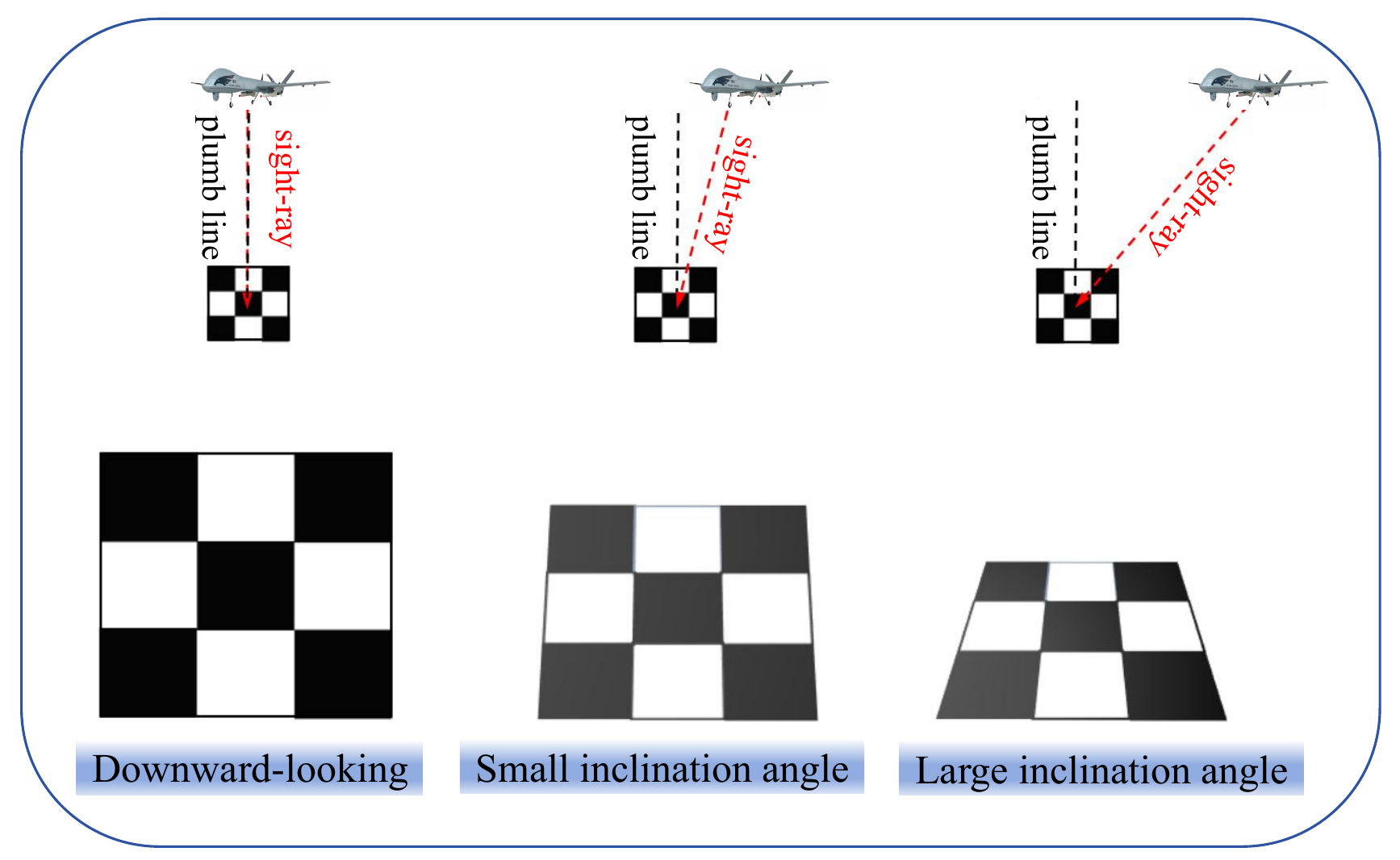}
\caption{Imaging maps of different inclination angles.}
\label{fig4}
\end{figure}

As proved in Section \ref{sec2.1}, when the target's motion is represented as temporal polynomials and the number of observations $N$ satisfies condition $2N \geq 3(K+1)$, there is a least squares solution for the target trajectory. As the number of observations increases, the reconstructed trajectory approaches the ground truth of the target trajectory. However, under limited observation conditions such as insufficient observations, long distance, and high observation error of platform, the 3D reconstruction error of trajectory intersection can be substantial or even cause degeneracies to occur. Long observation distance is a common limited observation condition in practical applications. Measurement errors caused by camera pose errors tend to amplify as the distance increases. Moreover, as shown in Figure \ref{fig4}, long-distance observation can also result in a large inclination angle.

Generally, the downward-looking is the optimal observation. However, in complex three-dimensional environments, observations of distant target areas often need to be conducted at a certain altitude. This will lead to the inclination angle. As the observation distance increases, the observation inclination angle becomes larger. As shown in Figure \ref{fig4}, under a large inclination angle, the geometric shape of the image undergoes significant distortion, and the angular error of the sight-ray increases notably. Therefore, the limited observations can cause severe ill-conditioning to the least squares system, i.e., small observation noise can lead to a significant measurement error. The least squares estimation results are unstable.

To improve the stability of target trajectory reconstruction under limited observation conditions, we introduce ridge estimation. By adding a penalty function to the least squares estimation equations, the proposed method mitigates the ill-conditioning, thereby improving the accuracy and stability of the system.

From Eq. (\ref{eq4}), the position of the target can be expressed as 
\begin{equation}
    \mathbf{P}_{i}=\tilde{\boldsymbol{\beta}}\tilde{\mathbf{\Theta}}_i,
    \label{eq6}
\end{equation} 
where 
\begin{equation}
    \tilde{\boldsymbol{\beta}}=\begin{bmatrix}a_0&a_1&\cdots&a_K\\b_0&b_1&\cdots&b_K\\c_0&c_1&\cdots&c_K\end{bmatrix}=\begin{bmatrix}\beta_a\\\beta_b\\\beta_c\end{bmatrix},
    \label{eq7}
\end{equation}

\begin{equation}
   \tilde{\mathbf{\Theta}}_i=\begin{bmatrix}t_i^0&t_i^1&\cdots&t_i^K\end{bmatrix}^\mathrm{T}.
   \label{eq8}
\end{equation}

Further, $\mathbf{P}_{i}$ can be rewritten as 
\begin{equation}
    \mathbf{P}_{i}=\mathbf{\Theta}_i\boldsymbol{\beta},
    \label{eq9}
\end{equation}
where 
\begin{equation}
    \mathbf{\Theta}_i=\mathbf{I}_{3\times3}\otimes\tilde{{\mathbf{\Theta}}}_i^\mathrm{T},
    \label{eq10}
\end{equation}

\begin{equation}
\boldsymbol{\beta}=\begin{bmatrix}\beta_a&\beta_b&\beta_c\end{bmatrix}^\mathrm{T},
\label{eq11}
\end{equation}
where $\otimes$ represents Kronecker Product. Let \(\mathbf{V}_i=(\mathbf{I}-\mathbf{l}_i\mathbf{l}_i^\mathrm{T})\), \(\mathbf{B}_i=(\mathbf{I}-\mathbf{l}_i\mathbf{l}_i^\mathrm{T})\mathbf{C_i}\) and \(\mathbf{A}_i=\mathbf{V}_i\mathbf{\Theta}_i\). Eq. (\ref{eq5}) can be written as a set of linear equations related to the target's motion parameters $\boldsymbol{\beta}$ : 
\begin{equation}
    \mathbf{A}_{i}\boldsymbol{\beta}=\mathbf{B}_{i},
    \label{eq12}
\end{equation} 
where only $\boldsymbol{\beta}$ is unknown. Combining the linear equations in Eq. (\ref{eq12}) of $N$ observations simultaneously, we can obtain the linear equations group:  
\begin{equation}
    \mathbf{A}\boldsymbol{\beta}=\mathbf{B},
    \label{eq13}
\end{equation} 
where \(\mathbf{A}\in\mathbb{R}^{3N\times3(K+1)}\), \(\mathbf{B}\in\mathbb{R}^{3N\times1}\). When $2N \geq 3(K+1)$, the least squares solution of the equation system Eq. (\ref{eq13}) is: 
\begin{equation}
 \boldsymbol{\beta}=(\mathbf{A}^\mathrm{T}\mathbf{A})^{-1}\mathbf{A}^\mathrm{T}\mathbf{B}.
 \label{eq14}
\end{equation}

However, under limited observation conditions, the 3D reconstruction error of the least squares estimation can be substantial or even cause degeneracies to occur. In such scenarios, the correlation of the observational data becomes significant, resulting in ill-conditioning of the least squares system. This ill-conditioning severely undermines the accuracy and robustness of the estimation results. To address this issue and enhance the accuracy and stability of reconstruction results under limited observation conditions, we introduce ridge estimation. 

Ridge estimation is an improved least squares estimation method. It improves the stability of the estimated result by sacrificing some precision to reduce the mean squared error while giving up the unbiasedness of least squares estimation. Ridge estimation is widely applied in data analysis across fields such as economics, engineering, and biomedicine. It performs very well in handling data with multicollinearity. Given the ill-conditioning problems, this paper introduces ridge estimation and adds a penalty function to the least squares estimation objective function. By appropriately reducing the accuracy, and sacrificing the unbiasedness of least squares, more robust and more realistic regression coefficients are obtained to solve the problem of serious multicollinearity of design matrix column vectors.

After adding the regularization term to the diagonal of the normal matrix of Eq. (\ref{eq14}), the ridge estimation can be expressed as:
\begin{equation}
    \boldsymbol{\beta}=(\mathbf{A}^\mathrm{T}\mathbf{A}-{r}\mathbf{I})^{-1}\mathbf{A}^\mathrm{T}\mathbf{B},
    \label{eq15}
\end{equation}
where $r$ is the ridge parameter, and \(\mathbf{I}\in\mathbb{R}^{3(K+1)\times3(K+1)}\) is the identity matrix. The ridge estimation is a linear transformation of the least squares estimation. Under limited observation conditions, there exists a $r>0$ such that the ridge estimation is better than the least squares estimation.

The selection of the ridge parameter $r$ is crucial for ridge estimation. Various methods exist to select the ridge parameters \cite{Golub1979,Lee1985}. In this paper, to obtain results quickly, we adopt the Hoerl-Kennard-Baldwin ridge parameter selection method \cite{Hoerl1975}. It is an efficient approach that does not require fitting or iteration. In practice, this method demonstrates good performance. We choose this method after considering both computational efficiency and accuracy. The calculation of the ridge parameter is as follows:
\begin{equation}
    r=\frac{t\delta_0^2}{\hat{\boldsymbol{\beta}}^\mathrm{T}\mathbf{A}^\mathrm{T}\mathbf{A}\hat{\boldsymbol{\beta}}},
    \label{eq16}
\end{equation}
where $t$ is the rank of matrix $\mathbf{A}$. When the number of observations $N$ satisfies $2N \geq 3(K+1)$, $t=3(K+1)$. $\hat{\boldsymbol{\beta}}$ is the result of the target's motion parameter calculated by the least squares estimate, and $\delta_0^2$ is calculated by:
\begin{equation}
    \delta_0^2=\frac{\mathbf{B}^\mathrm{T}\left[\mathbf{I}-\mathbf{A}(\mathbf{A}^\mathrm{T}\mathbf{A})^{-1}\mathbf{A}^\mathrm{T}\right]\mathbf{B}}{n-t}.
    \label{eq17}
\end{equation}

\subsection{Automatic selection method for the order parameter}\label{sec2.3}

Our approach requires selecting the order of the temporal polynomials, $K$. This parameter was manually adjusted in previous trajectory intersection methods based on experience. The order of the temporal polynomials controls the complexity of the reconstructed trajectory and the number of parameters to be estimated. Selecting the order of the temporal polynomials correctly is crucial to the accuracy of the 3D reconstruction of the target trajectory. If the value of the selected $K$ is excessively high, the algorithm tends to overfit the measurement noise. On the other hand, if the value is too low, the trajectory reconstructed by the algorithm fails to capture the intricacies of the target's motion. In this section, we present an algorithm to automatically determine the optimal value of $K$ rather than manually setting a value.

To automatically select the order of the temporal polynomials, we construct an objective function based on geometric error. Specifically, for each observation time $t_j$, we compute the direction vector from the optical center of the camera to the reconstructed target, denoted as $\hat{\mathbf{l}}_j$. Subsequently, for each value $K_i$, we calculate the sum of the squared sight-ray errors at all $t_j$. Then, we construct the following objective function:
\begin{equation}
    K_i^*=\underset{K_i}{\text{argmin}}\sum_{j=1}^N\lVert \hat{\mathbf{l}}_j-{\mathbf{l}}_j \rVert_2,
    \label{eq18}
\end{equation}
where $K_i=0,1,2,3$. Because within a certain period of time, the ground target's motion usually follows certain physical laws, such as static state, uniform linear motion, or uniform accelerated motion. Taking $K_i$ as 0, 1, 2, or 3 can accurately reconstruct the target trajectory and achieve high efficiency simultaneously. It is worth mentioning that in the method of reference \cite{Park2015}, the automatic selection of the number of DCT trajectory basis vectors $K^{DCT}$ requires calculating the reprojection error for \(K_i^{CDT}=1,2,\ldots,\lfloor2F/3\rfloor \), where $\lfloor\cdot\rfloor$  is the floor operator (the largest integer not greater than $\cdot$ ). Compared to our method, the reference method \cite{Park2015} requires a significant amount of time to automatically select the value of $K$. It is an advantage of temporal polynomials that they carry significant physical meaning. When $K_i$ is high, the trajectory tends to fit the measurement noise excessively, leading to an increased reprojection error for the sight-rays. In contrast, when $K_i$ is low, the reprojection error for the sight-rays remains high because of the restricted expressiveness of the temporal polynomials. Therefore, when minimizing the objective function in Eq. (\ref{eq18}), we only need four iterations to determine the optimal order of temporal polynomials.

\subsection{3D trajectory reconstructability analysis}\label{sec2.4}

\begin{figure}[htbp] 
\centering
\subfloat[]{\includegraphics[width=2.1in]{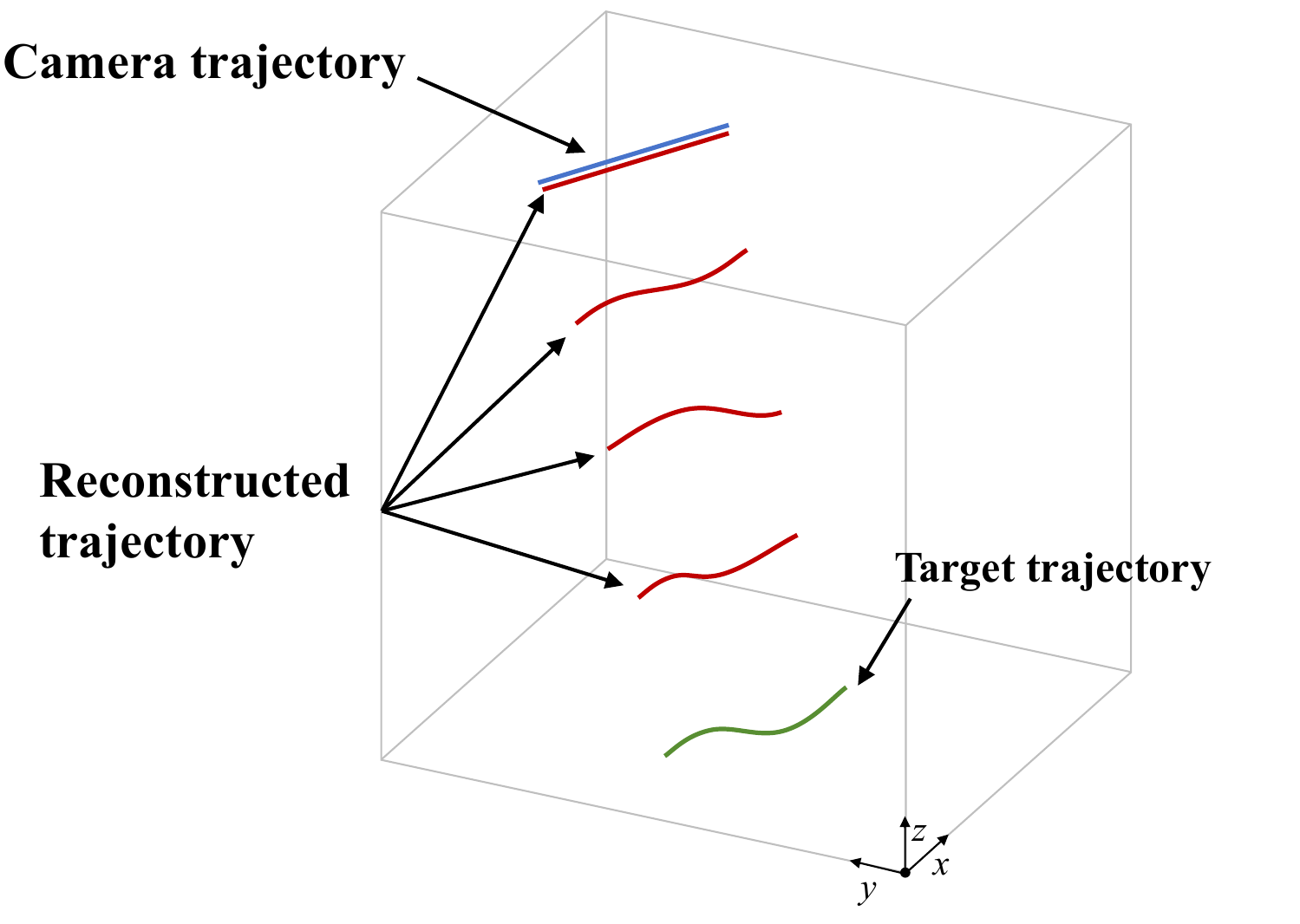}%
\label{fig5a}}
\hfil
\subfloat[]{\includegraphics[width=2.1in]{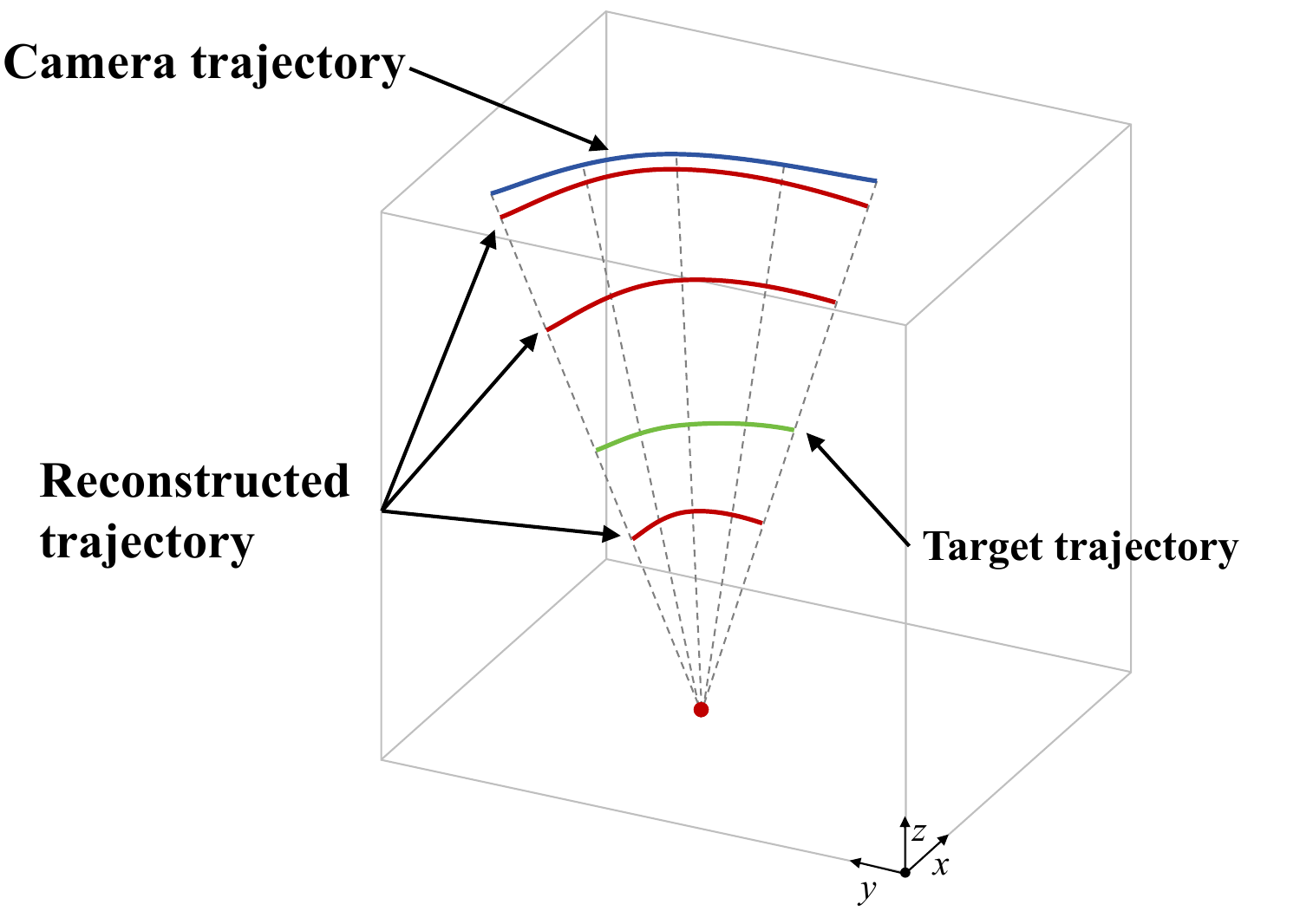}%
\label{fig5b}}
\hfil
\subfloat[]{\includegraphics[width=2.1in]{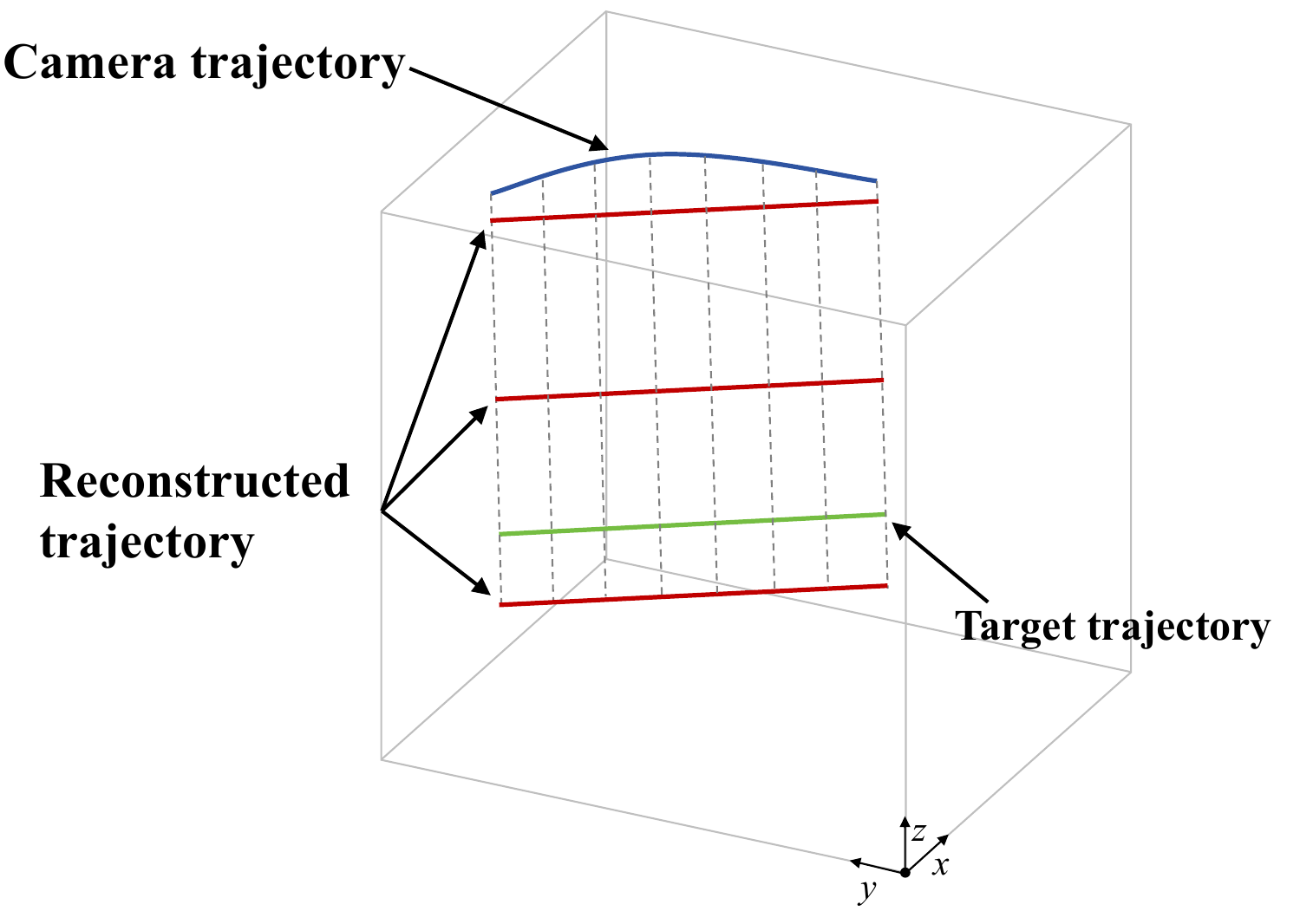}%
\label{fig5c}}
\caption{Degeneracy situations. (a) The order of the temporal polynomial representing the camera motion is lower than that of the target's motion. (b) All sight-rays intersect at the same point. (c) All sight-rays are parallel.}
\label{fig5}
\end{figure}

This section performs a geometric analysis of the camera motion, target motion, and temporal polynomials in our algorithm. As previous work proved \cite{Yu2009,Zhou2015}, there are two situations where a definite solution cannot be obtained, as shown in Figure \ref{fig5}: 

(i) The order of the temporal polynomials representing the camera motion is equal to or lower than that of the target motion, as shown in Figure \ref{fig5a}. 

(ii) All sight-rays intersect at the same point (all sight-rays being parallel is a special case of this situation, where they intersect at the infinite point), as shown in Figure \ref{fig5b} and Figure \ref{fig5c}.

However, in practical applications, it is noted that the reconstructed trajectories of the target closely align with the ground truth when the camera motion relative to the target motion is sufficiently significant in practice. Conversely, if the camera motion is simple compared to the target motion, the solution is likely to deviate from the ground truth. Park et al.\cite{Park2015} proposed an index called \textit{reconstructability} for DCT trajectory basis vectors to measure the reconstruction accuracy of solvable systems. The \textit{reconstructability} is extended from DCT trajectory basis vectors to temporal polynomials by analyzing the geometric relationship among camera motion, target motion, and temporal polynomials. The definition of \textit{reconstructability} enables us to quantify the reconstruction accuracy. Why the estimated trajectory degenerates to be close to the camera trajectory can be explained using this definition. The \textit{reconstructability} based on temporal polynomials is defined as follows.

\begin{figure}[htbp]
\centering
\includegraphics[width=3in]{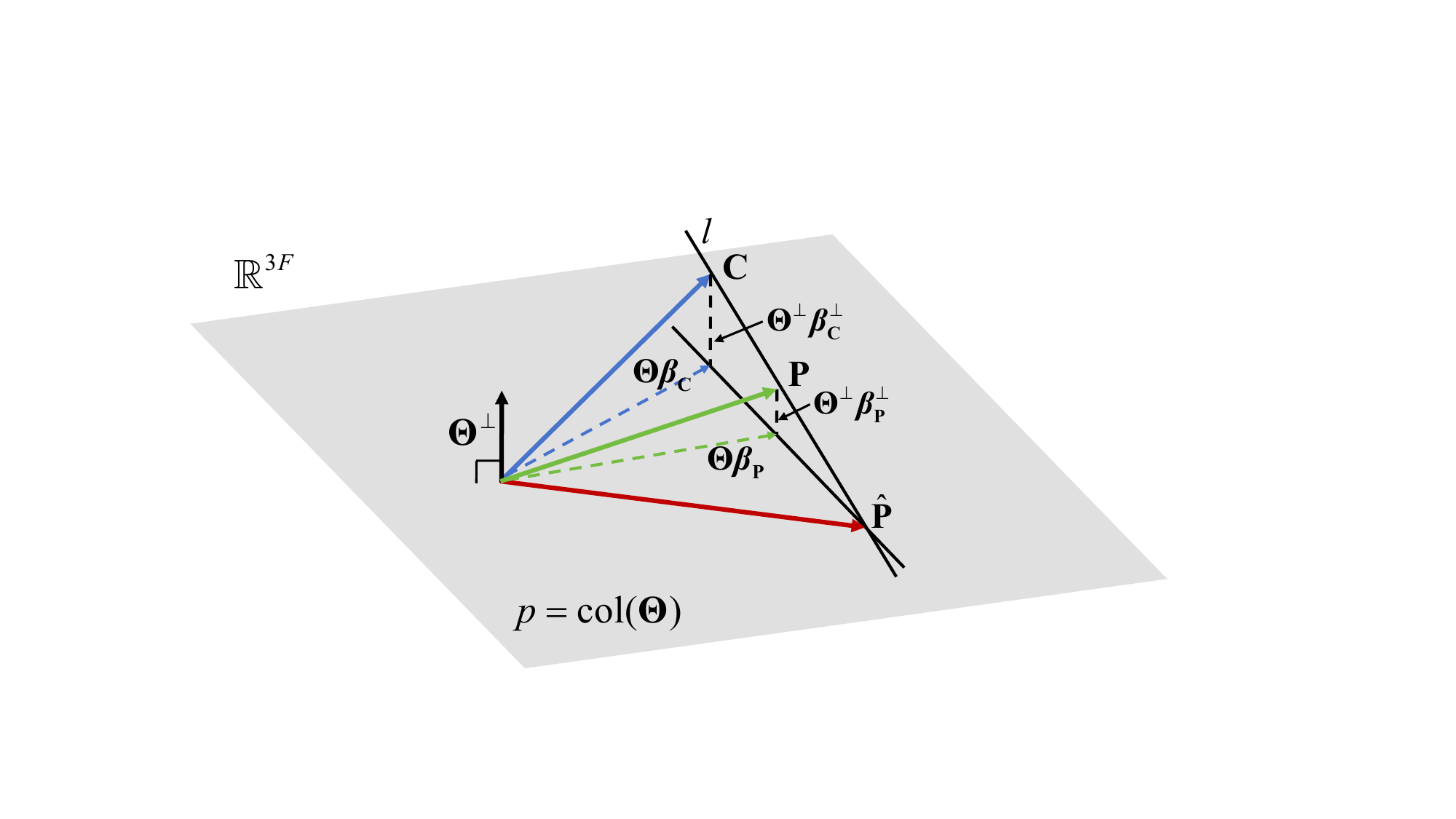}
\caption{Geometric illustration of the least squares solution, when the order of the temporal polynomials representing the camera motion is higher than that of the target's motion.}
\label{fig6}
\end{figure}

As shown in Figure \ref{fig6}, let $\mathbf{C}$ and $\mathbf{P}$ be the trajectory of the camera and the ground truth of the target trajectory represented by temporal polynomials, respectively. The plane $p=\operatorname{col}(\mathbf{\Theta})$ is the subspace spanned by different orders of time $t$. Thus $\mathrm{col}(\mathbf{\Theta}^{\perp})$ represents the null space of the matrix $\mathbf{\Theta}$. It can be seen from Figure \ref{fig6} that the estimated position of the target lies in plane $p$ and the line $l$ that connects $\mathbf{C}$ and $\mathbf{P}$, simultaneously. Note that the line and the plane are a conceptual 3D vector space representation for the 3$N$-dimensional space. Let the trajectory of the camera $\mathbf{C}$ and the ground truth of the target $\mathbf{P}$ be projected onto the plane $p$ as $\mathbf{\Theta}{\boldsymbol{\beta}}_\mathbf{C}$ and $\mathbf{\Theta}{\boldsymbol{\beta}}_\mathbf{P}$, respectively. Then $\mathbf{C}$ and $\mathbf{P}$ can be decomposed into the column space of $\mathbf{\Theta}$ and that of the null space, $\mathbf{\Theta}^{\perp}$ as follows: 
\begin{equation}
\mathbf{C}=\mathbf{\Theta}\boldsymbol{\beta}_{\mathbf{C}}+\mathbf{\Theta}^{\perp}\boldsymbol{\beta}_{\mathbf{C}}^{\perp}
\label{eq19}
\end{equation}
\begin{equation}\mathbf{P}=\mathbf{\Theta}\boldsymbol{\beta}_{\mathbf{P}}+\mathbf{\Theta}^{\perp}\boldsymbol{\beta}_{\mathbf{P}}^{\perp}
\label{eq20}
\end{equation}
where $\boldsymbol{\beta}^{\perp}$ is the coefficient vector for the null space. \(\mathbf{\Theta}^{\perp}\boldsymbol{\beta}_{\mathbf{C}}^{\perp}\) and \(\mathbf{\Theta}^{\perp}\boldsymbol{\beta}_{\mathbf{P}}^{\perp}\) are the component of the null space  $\mathbf{\Theta}^{\perp}$, i.e. the part that cannot be represented by the temporal polynomials. Thus, the \textit{reconstructability} $\eta$ of the target trajectory for temporal polynomials can be defined as:
\begin{equation}
\eta\left(\boldsymbol{\Theta}\right)=\frac{\left\|\boldsymbol{\Theta}^\perp\boldsymbol{\beta}_\mathbf{C}^\perp\right\|}{\left\|\boldsymbol{\Theta}^\perp\boldsymbol{\beta}_\mathbf{P}^\perp\right\|}\simeq\frac{\text{How poorly }\boldsymbol{\Theta} \text{ describes }\mathbf{C}}{\text{How poorly }\boldsymbol{\Theta}\text{ describes }\mathbf{P}}.
\label{eq21}
\end{equation}

From Eq. (\ref{eq21}) and Figure \ref{fig6}, it can be seen that the larger the value of $\eta$, the higher the reconstruction accuracy of the target trajectory. And when \(\eta\rightarrow\infty\), the estimated trajectory is equal to the ground truth. This shows the importance of the correct selection of the temporal polynomials' order $K$.

\begin{figure}[htbp] 
\centering
\subfloat[]{\includegraphics[width=3in]{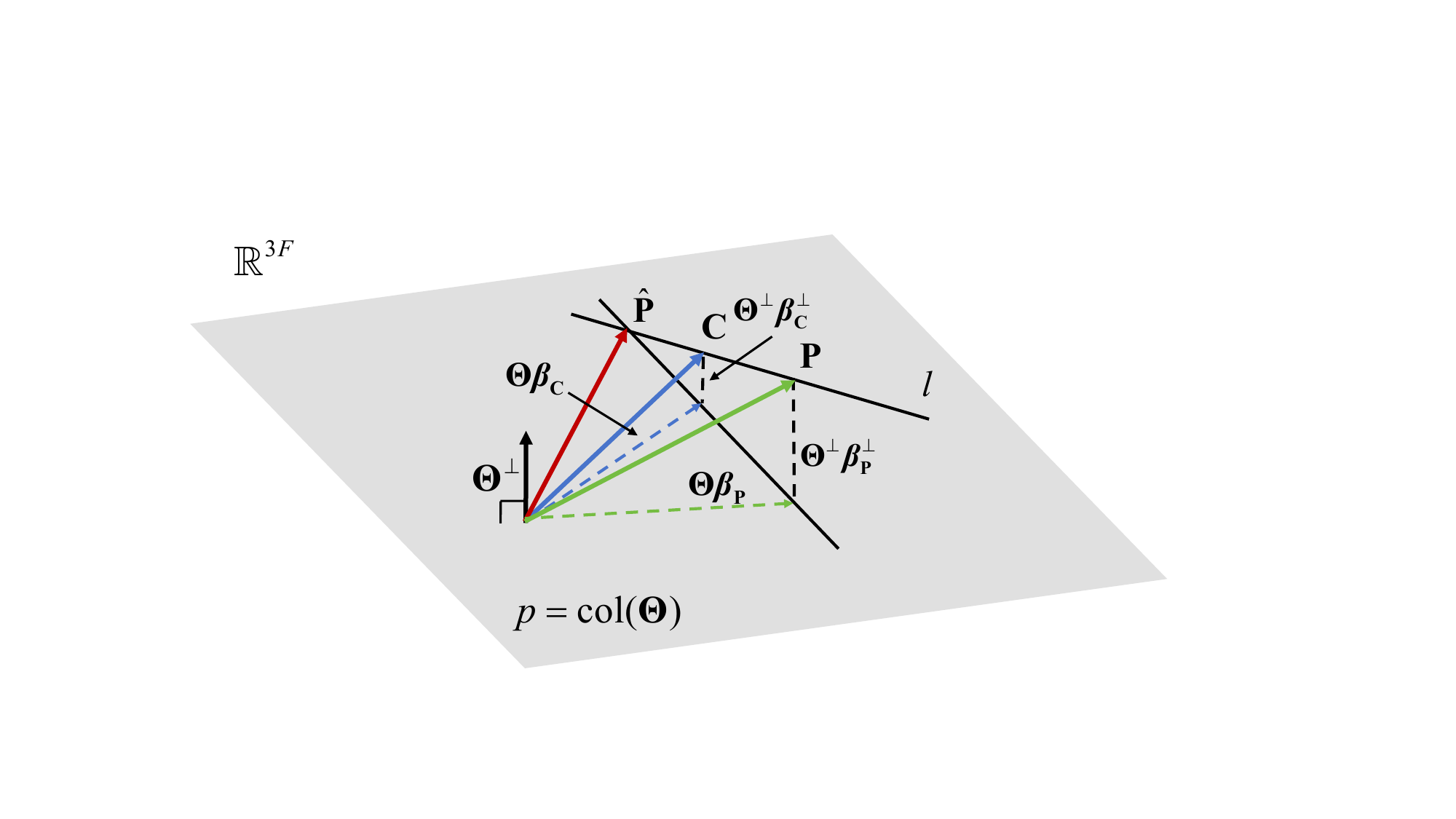}%
\label{fig7a}}
\hfil
\subfloat[]{\includegraphics[width=3in]{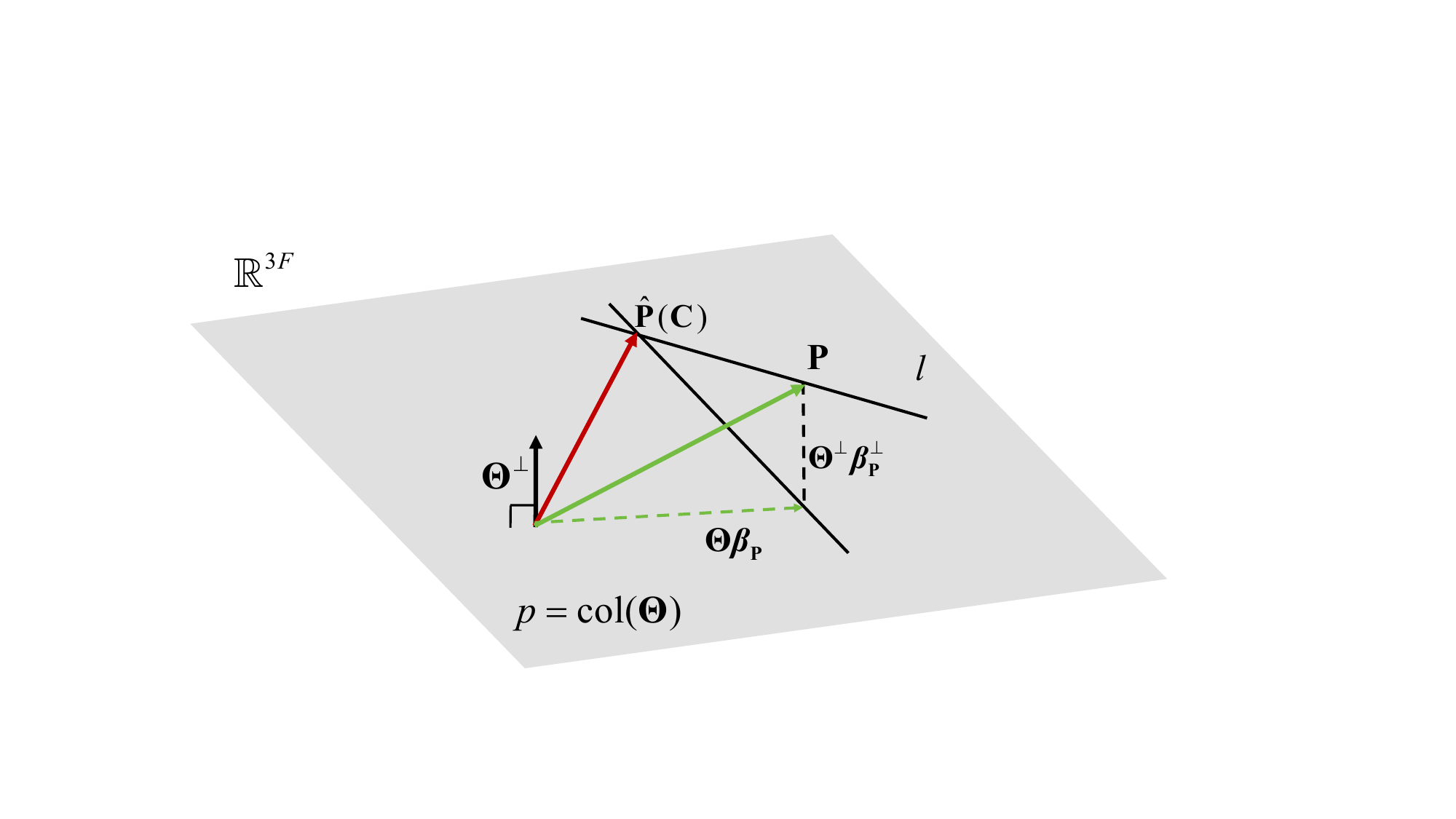}%
\label{fig7b}}
\caption{Geometric illustration of degeneracy situations. (a) The order of the temporal polynomials representing the camera motion is lower than that of the target's motion. (b) The order of the temporal polynomials representing the camera motion is consistent with the selected order.}
\label{fig7}
\end{figure}

The reconstruction accuracy can be described quantitatively based on the \textit{reconstructability}. A more detailed geometric analysis of the measurement system can also be conducted. For example, from Eq. (\ref{eq21}) and Figure \ref{fig6}, the first situation of unsolvable systems that were proposed in previous work can be more intuitively proved. Moreover, why the estimated trajectory degenerates to be close to the camera trajectory in practice can be explained. As shown in Figure \ref{fig7a}, when the order of the temporal polynomials representing the camera motion is lower than that of the target's motion, we can have \(\left\|\boldsymbol{\Theta}^\perp\boldsymbol{\beta}_\mathbf{C}^\perp\right\|<\left\|\boldsymbol{\Theta}^\perp\boldsymbol{\beta}_\mathbf{P}^\perp\right\| \). Thus, the intersection of line $l$ and plane $p$ is closer to $\mathbf{C}$ but far away from $\mathbf{P}$ and this leads to incorrect estimation result $\hat{\mathbf{P}}$. As shown in Figure \ref{fig7b}, when the selected order parameter $K$ can well describe the camera trajectory $\mathbf{C}$, the estimated result $\hat{\mathbf{P}}$ is identical to the camera trajectory. This can explain why the estimated trajectory degenerates to be close to the camera trajectory in practice when the camera motion is relatively simple. 

\begin{figure}[htbp]
\centering
\includegraphics[width=3in]{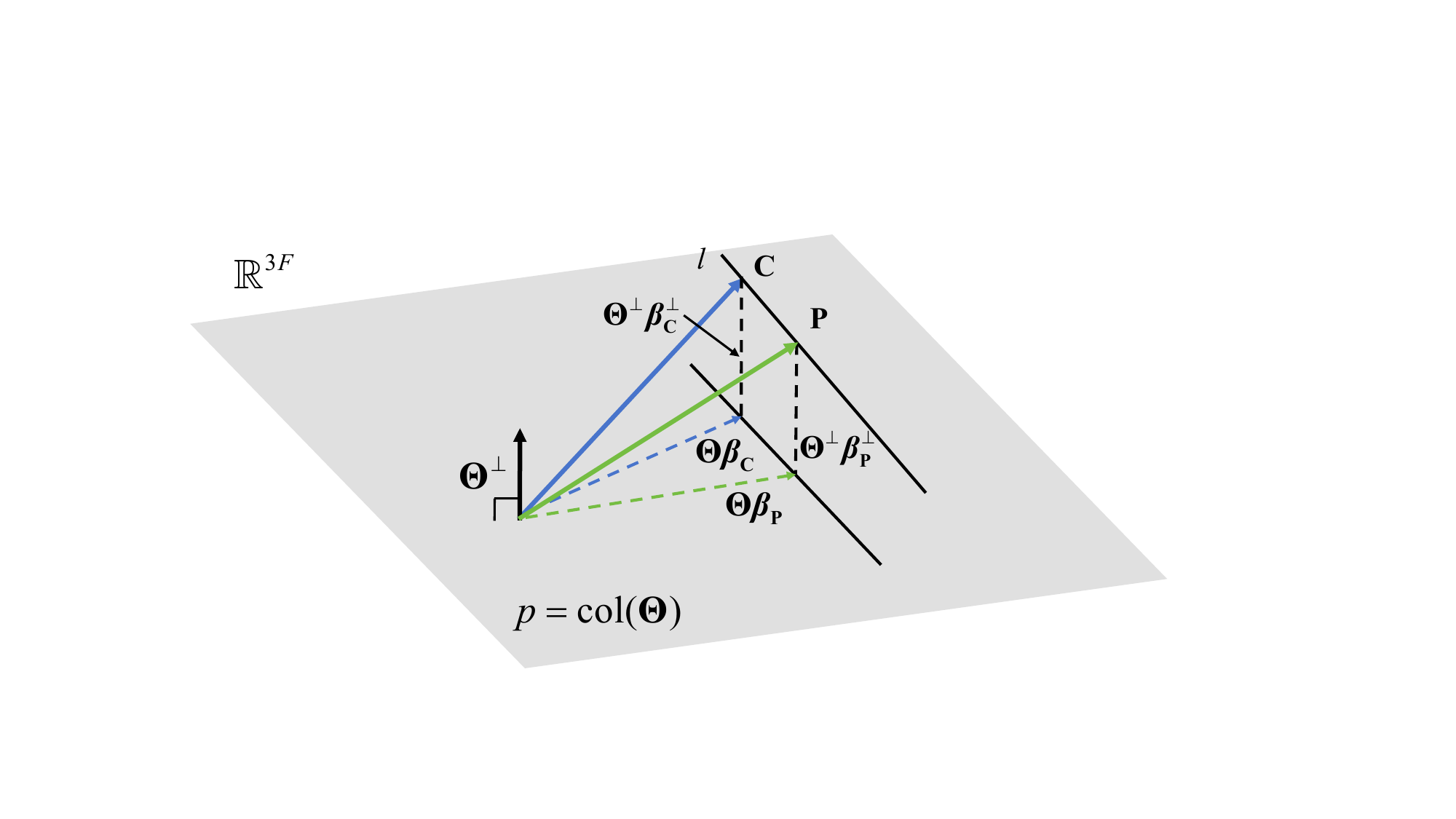}
\caption{Geometric illustration of a kind of limited observation condition.}
\label{fig8}
\end{figure}

Another kind of limited observation condition can also be analyzed using the definition of \textit{reconstructability}. As shown in Figure \ref{fig8}, the complexity of the camera trajectory is similar to that of the target trajectory, i.e.,  \(\left\|\boldsymbol{\Theta}^\perp\boldsymbol{\beta}_\mathbf{C}^\perp\right\|\approx\left\|\boldsymbol{\Theta}^\perp\boldsymbol{\beta}_\mathbf{P}^\perp\right\| \). If there is no noise, the camera trajectory is more complex, and the chosen order of temporal polynomial $K$ is appropriate, the target trajectory can theoretically be solved. However, in practical applications, the estimation results often degrade due to the influence of system noise. As shown in Figure \ref{fig8}, the measurement error can be significant under such conditions. Under such conditions, a small observation error can lead to substantial measurement errors, or even cause degradation. Therefore, such limited observation conditions can also cause ill-conditioning to the least squares solution system. Ridge estimation can be introduced to mitigate the ill-conditioning.

Through the analysis of \textit{reconstructability}, the geometric relationship among camera trajectory, target trajectory, and temporal polynomials have been more intuitively determined. We have also analyzed another kind of limited observation condition. 
\section{Simulated experiments}\label{sec3}

In this section, we evaluate the performance of the proposed method on simulated data. Section \ref{sec3.1} conducts simulated experiments under a low noise level to evaluate the feasibility of the three general types of methods for the issues studied in this paper. The experimental results demonstrate the feasibility of using temporal polynomials to describe the target trajectory. Further, we analyze the reasons for the inapplicability of the other two types of methods. Section \ref{sec3.2} evaluates the accuracy and robustness of the proposed method on the simulated data under limited observation conditions. Section \ref{sec3.3} evaluates the performance of the proposed $K$ selection algorithm. Section \ref{sec3.4} evaluates the performance of the proposed method under missing data situations.

\subsection{Feasibility}\label{sec3.1}

In this section, we evaluate the feasibility of the three general types of methods using simulated data. This paper primarily addresses the 3D trajectory reconstruction of moving targets such as vehicles and ships and aims to provide precise 3D trajectory reconstruction results in a short time. There are three general types of methods based on motion assumptions of the moving points. The previous works of trajectory intersection methods have proven that they can effectively measure the trajectory of vehicles or ships using a monocular camera by representing the target trajectory as temporal polynomials \cite{Yu2009,Li2014,Chen2019}. However, when solving these motions, the reconstruction accuracy and computational efficiency of the trajectory triangulation methods and the DCT trajectory basis vectors methods are inadequate.

\begin{figure}[htbp]
\centering
\includegraphics[width=2.5in]{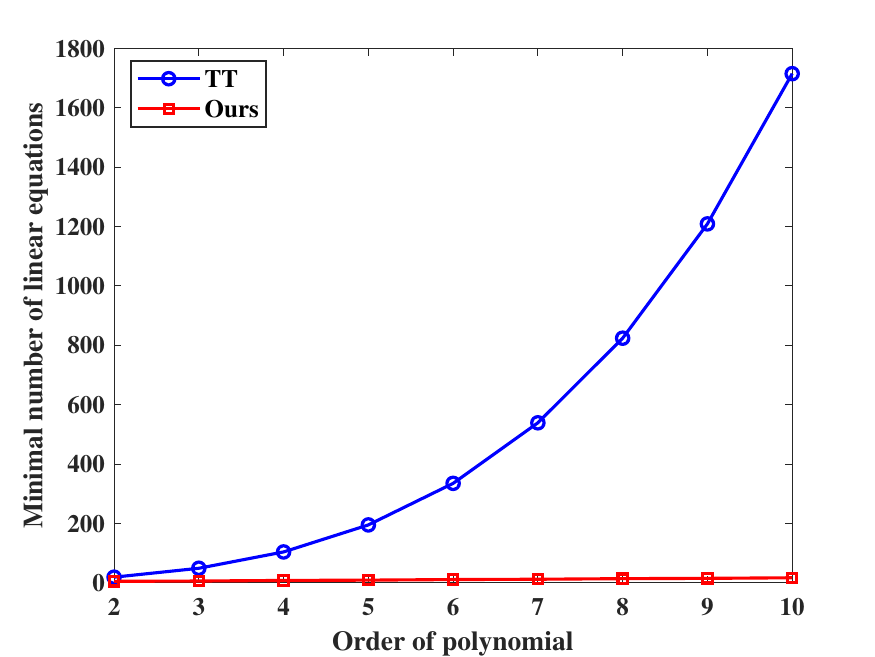}
\caption{Minimal number of linear equations need to be solved in different orders of polynomial.}
\label{fig9}
\end{figure}

Although the state-of-the-art trajectory triangulation algorithm \cite{Kaminski2004} (referred to as TT) can utilize linear equations to reconstruct diverse trajectories expressible as polynomials, it requires many more equations at the same order of the polynomial. Under the same $K$ value, our algorithm requires at least $\lfloor\frac{3}{2}(K+1)\rfloor$ equations. However, the TT algorithm requires solving at least \(N_{d}=\begin{pmatrix}d+5\\d\end{pmatrix}-\begin{pmatrix}d+3\\d-2\end{pmatrix}-1\) equations. where \(\begin{pmatrix}\cdot\\\cdot\end{pmatrix}\) represents the combinatorial number. Therefore, the minimal number of linear equations of the TT algorithm increases exponentially while that of our algorithm increases linearly, as shown in Figure \ref{fig9}. Some TT methods even require the optimization of nonlinear equations and can only reconstruct the targets moving along a line \cite{Avidan1999,Avidan2000}. These methods acquire a good initial value. Thus, the TT methods are not suitable for accurately recovering the 3D trajectory of the target within a short period of time.

As for the DCT trajectory basis vectors based method \cite{Park2015} (referred to as DCT), theoretically, the DCT trajectory basis vectors can represent any object trajectory without prior information \cite{Akhter2011}. Thus, this method can reconstruct arbitrary trajectories. However, considering the problem addressed in this paper, for vehicle and ship targets, the most common motion patterns are uniform linear motion, and uniform accelerated motion. Referring to the motion patterns of vehicles and ships, the simulated motions of the point targets are set as: 
\[\begin{cases}X=10+5t\\Y=5t\\Z=t\end{cases},
\begin{cases}X=10+t^2\\Y=13+2t^2\\Z=0.5t^2\end{cases},\]
where the simulated moving points are in uniform linear motion and uniform accelerated motion, respectively. Let the observation platform equipped with a monocular camera perform circular motion to ensure a definite solution for the system. The trajectory of the camera's optical center is: 
\[  \begin{cases}X_C=100\sin(\frac{t}{10\pi})\\Y_C=100-100\cos(\frac{t}{10\pi})\\Z_C=100\end{cases}.\]

\begin{figure}[htbp]
\centering
\subfloat[]{\includegraphics[width=3in]{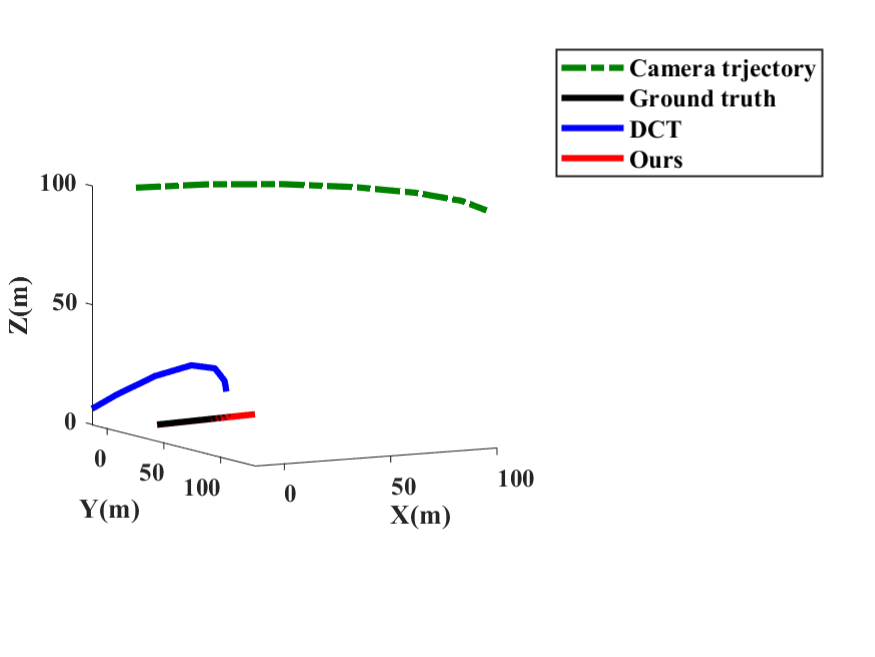}%
\label{fig10a}}
\hfil
\subfloat[]{\includegraphics[width=3in]{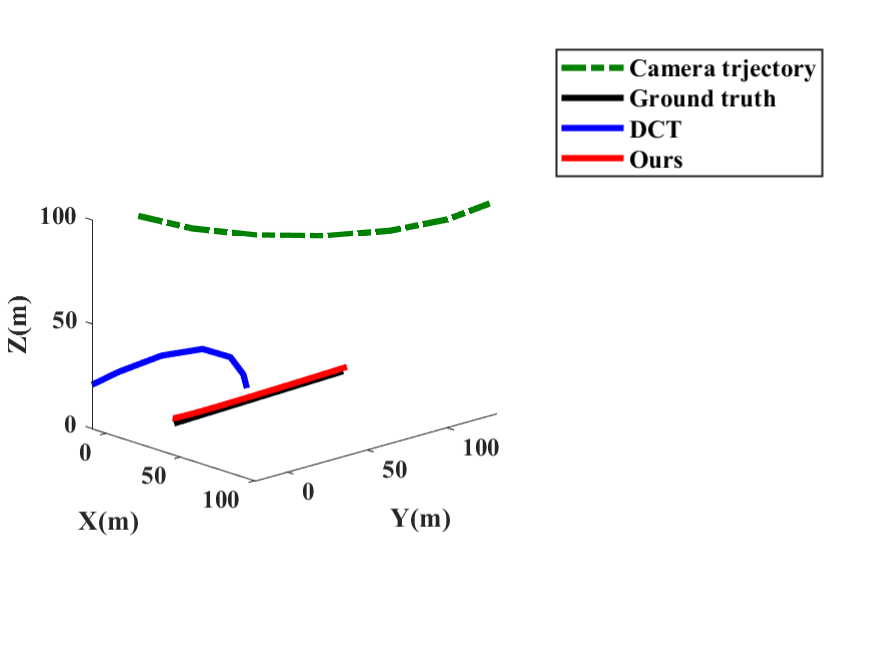}%
\label{fig10b}}
\caption{We evaluate the DCT method \cite{Park2015} and our algorithm on simulated data under a low noise level. (a) The result on uniform linear motion data. (b) The result on uniform accelerated motion data.}
\label{fig10}
\end{figure}

The trajectory of the camera's optical center and the direction of the sight-ray at each observation time are known. In the simulated data, a low level of system noise is introduced. The minimal solution of the proposed method only needs 3 observations for uniform linear motion and 5 for uniform accelerated motion. The total number of observations in the simulated data is 60, far greater than the minimum required observations. When reconstructing the trajectories of targets moving in uniform linear motion and uniform accelerated motion using the DCT method, there is a significant degeneracy even when the number of observations reaches 60. As shown in Figure \ref{fig10}, the estimated trajectories by the DCT method significantly deviate from the ground truths. However, our algorithm can accurately reconstruct the target trajectory under a low noise level and sufficient observation conditions. This may be because while DCT trajectory basis vectors can easily represent arbitrarily complex motions, simple motions like uniform linear motion require a large number of DCT trajectory basis vectors to represent them. When solving simple motions, DCT method tends to overfit the measurement noise. Thus, the DCT method exhibits significant degeneracy. The DCT method is more applicable to human motion but is unsuitable for reconstructing vehicle and ship trajectories. For the problem researched in this paper, representing the target trajectory as temporal polynomials is the optimal method.

\subsection{Accuracy and robustness}\label{sec3.2}

In this section, we evaluate the accuracy and robustness of the proposed method on the simulated data. The previous works of trajectory intersection methods have proven that they can effectively measure the trajectory of vehicles, ships, and other moving targets using a monocular camera by representing the target trajectory as temporal polynomials \cite{Yu2009,Li2014,Chen2019}. However, under limited observation conditions such as insufficient observations, long distance, high observation error of platform, and low motion complexity of the observation platform, the previous methods are unstable. They can even cause degeneracies because of the severe ill-conditioning of the least squares equation system caused by limited observation conditions. Therefore, this paper introduces ridge estimation to the trajectory intersection method to mitigate the ill-conditioned problem and improve the stability under limited observation conditions.

\begin{figure}[htbp] 
\centering
\subfloat[]{\includegraphics[width=2.5in]{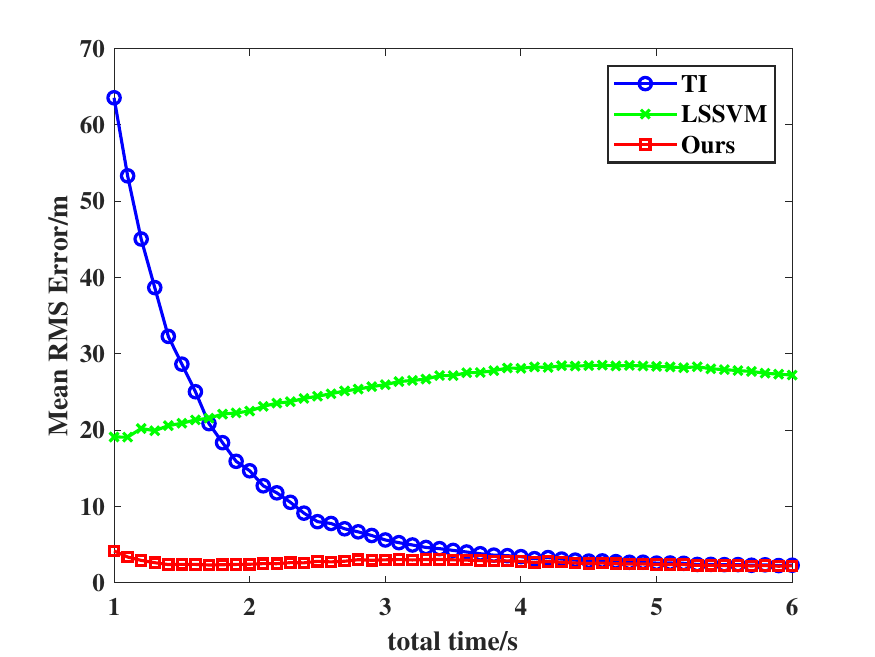}%
\label{fig11a}}
\hfil
\subfloat[]{\includegraphics[width=2.5in]{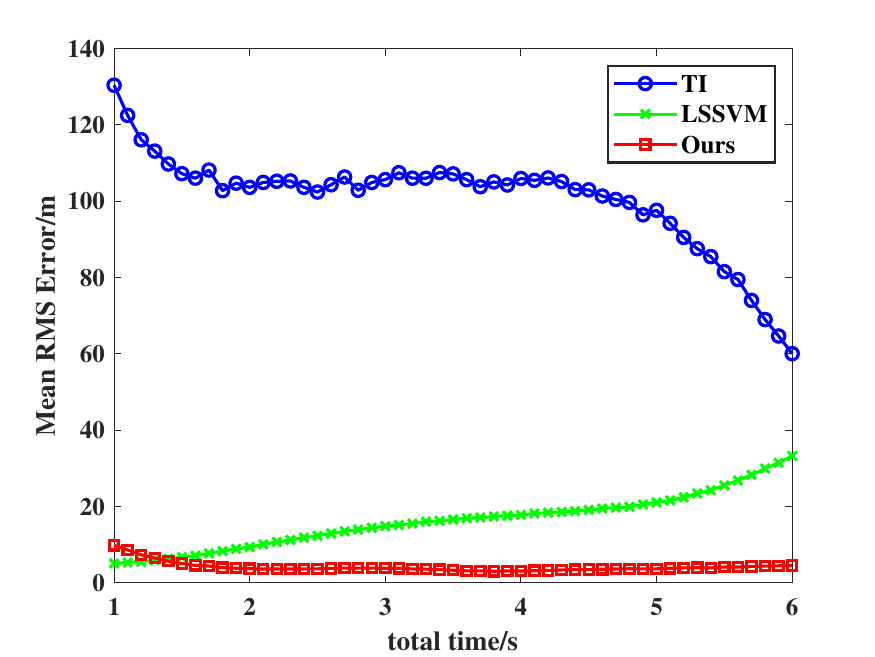}%
\label{fig11b}}
\caption{We evaluate the TI method \cite{Yu2009}, the LSSVM method \cite{Li2014}, and our algorithm on uniform linear motion and uniform accelerated motion data. (a) The result of uniform linear motion data. (b) The result of uniform accelerated motion data.}
\label{fig11}
\end{figure}

To quantitatively evaluate our 3D trajectory reconstruction method, the trajectory of the point targets and the camera’s optical center are set as Section \ref{sec3.1}. The target trajectory is estimated by the trajectory intersection method \cite{Yu2009} (referred to as TI), the LSSVM method \cite{Li2014}, and the proposed method, respectively. However, we introduce a high level of variety noises satisfying the normal distribution with a mean of zero, including the position systematic noise of the camera's optical center with a standard deviation of 1 m, the position random noise of the camera's optical center with a standard deviation of 1 m, the angle systematic noise of the sight-ray with a standard deviation of 0.3°, and the angle random noise of the sight-ray with a standard deviation of 0.3°. The frame rate is set at 10 Hz and 1000 independent experiments are conducted under the observation times from 1 to 6 s. Estimate the target trajectory using the TI method \cite{Yu2009}, the LSSVM method \cite{Li2014} and our algorithm respectively, and calculate the mean root mean square (RMS) error of the target position at each total observation time. The experimental results are shown in Figure \ref{fig11}. 

It can be seen from Figure \ref{fig11a}, when the target is in uniform linear motion, the LSSVM method exhibits higher accuracy when the number of observations is insufficient. However, when sufficient observations are available, the accuracy of the LSSVM method is lower. When the target is moving in uniform accelerated motion, the LSSVM method demonstrates higher accuracy, as shown in Figure \ref{fig11b}. This may be attributed to that the trajectory estimated by the LSSVM method does not strictly adhere to the determined order. For linear motion, it is more likely to cause significant errors. However, as shown in Figure \ref{fig11}, in the both two motion patterns of the targets, the proposed algorithm achieved the highest accuracy, especially under the ill-conditioning of a small number of observations. As shown in Figure \ref{fig11a}, when the number of observations is relatively low, the accuracy of our algorithm is significantly higher than that of the TI method. As the number of observations increases, the accuracy of both methods gradually approaches. However, although it appears to be very close in Figure \ref{fig11a}, our algorithm consistently outperforms the TI method.

\begin{figure}[htbp]  
\centering
\subfloat[]{\includegraphics[width=3in]{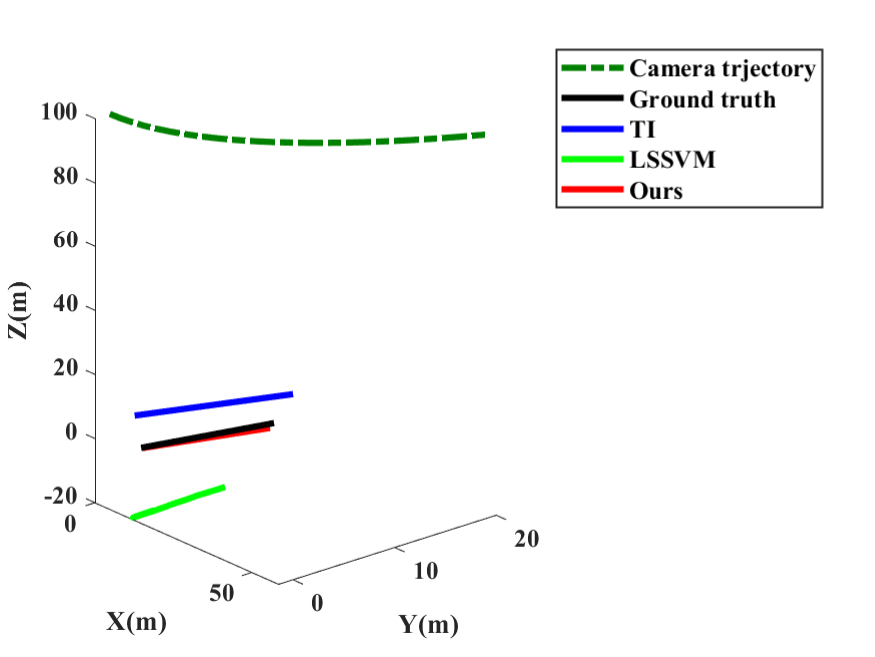}%
\label{fig12a}}
\hfil
\subfloat[]{\includegraphics[width=3in]{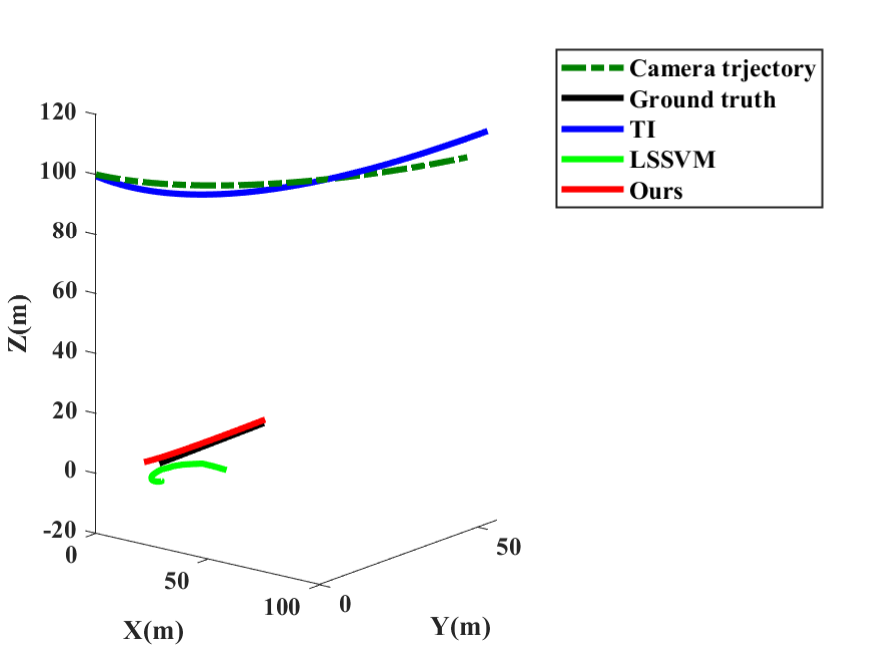}%
\label{fig12b}}
\caption{Illustration of experimental results under low \textit{reconstructability} condition. (a) The result of uniform linear motion data with a total time of 2 s. (b) The result of uniform accelerated motion data with a total time of 3.5 s.}
\label{fig12}
\end{figure}

Theoretically, to obtain a definite solution for the target trajectory, the number of observations $N$ only needs to satisfy the condition $2N \geq 3(K+1)$. However, due to various observation noises in practice, more observations are usually required to ensure the estimation accuracy. By introducing ridge estimation, our algorithm has significantly improved the estimation accuracy under fewer observations. Our algorithm only requires an observation time of 1 s to achieve almost the same estimation accuracy as the TI method with 3 s when the target moves in uniform linear motion, as Figure \ref{fig11a}. As shown in Figure \ref{fig11b}, when the total observation time is less than 5 s, the TI result occurs degeneracy, but our algorithm can still reconstruct the target trajectory accurately. Figure \ref{fig12a} shows the situation at the total observation time of 2 s in Figure \ref{fig11a} and Figure \ref{fig12b} shows the situation at the total time of 3.5 s in Figure \ref{fig11b}. The targets are in uniform linear motion and uniform accelerated motion, respectively. Figure \ref{fig12a} shows that the trajectories reconstructed by the TI and the LSSVM methods exhibit low degradation accuracy. However, our algorithm can accurately reconstruct the target trajectory. The average RMS errors for the TI, LSSVM, and our methods are 13.33 m, 21.47 m, and 2.46 m, respectively. It can be seen in Figure \ref{fig12b} that the trajectory reconstructed by the TI method degrades to be close to the camera trajectory as the proved situation in Section \ref{sec2.4}. Although the trajectory reconstructed by the LSSVM method does not degrade to be close to the camera trajectory as the TI method, its shape still exhibits degeneracy, showing significant differences from the ground truth. However, our algorithm can accurately reconstruct the target trajectory. The average RMS errors for the three methods are 106.89 m, 17.81 m, and 3.13 m, respectively. The simulated experimental results demonstrate that our algorithm can mitigate the ill-conditioned problem and improve the accuracy and robustness.

\subsection{Performance of the $K$ selection method}\label{sec3.3}

\begin{figure}[htbp]  
\centering
\subfloat[]{\includegraphics[width=2.5in]{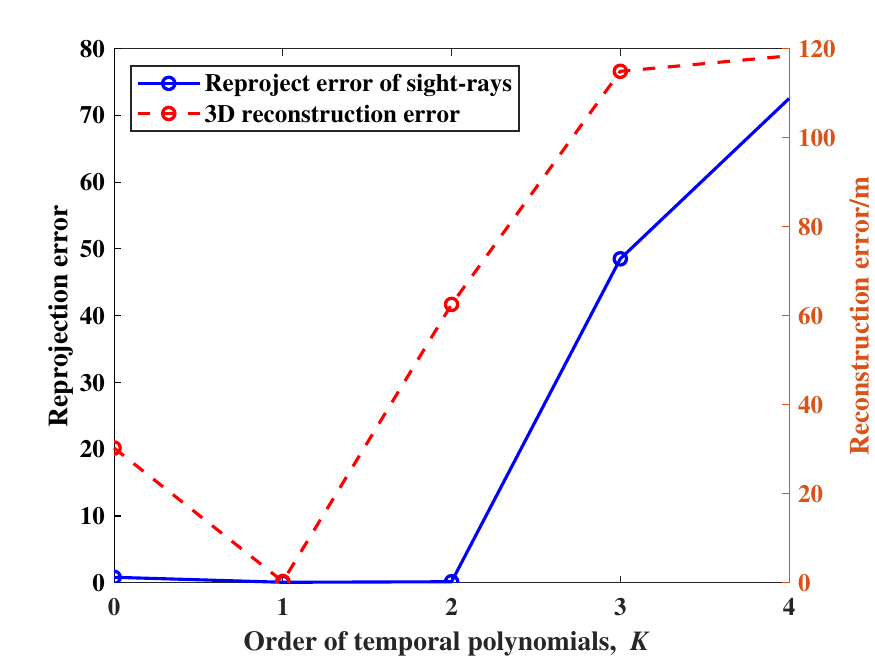}%
\label{fig13a}}
\hfil
\subfloat[]{\includegraphics[width=2.5in]{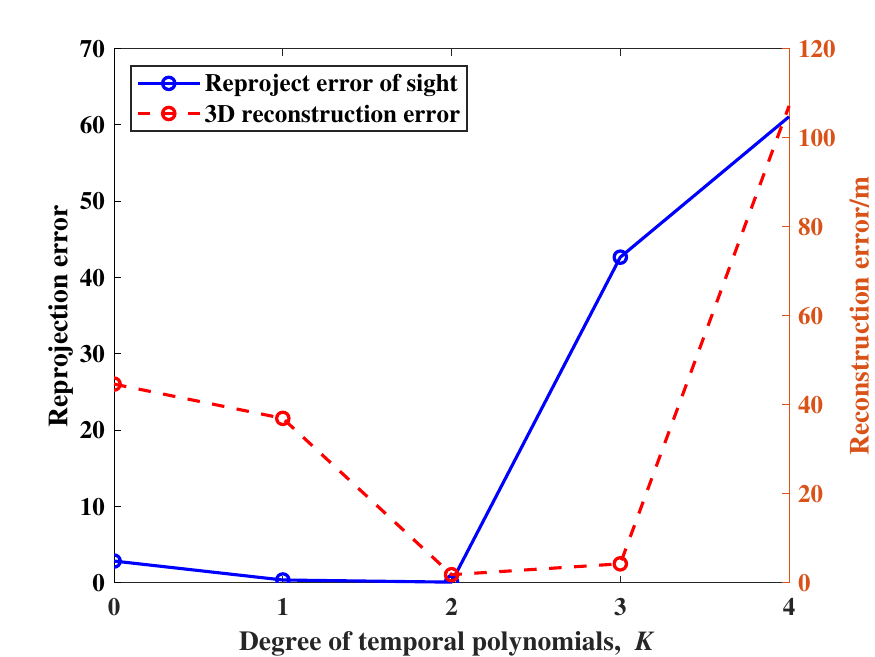}%
\label{fig13b}}
\caption{3D reconstruction errors and reprojection errors of sight-rays calculated using deferent orders of temporal polynomials, $K$. (a) The result of uniform linear motion. (b) The result of uniform accelerated motion.}
\label{fig13}
\end{figure}

In this section, we evaluate the performance of the proposed $K$ selection algorithm on simulated data. The target points are set to be moving in uniform linear motion and uniform accelerated motion. The corresponding $K$ values are 1 and 2. The camera's optical center is moving in a circular motion. For each motion pattern, various noises as Section \ref{sec3.2} are added. The method proposed in Section \ref{sec2.3} is used to select the value of $K$. As shown in Figure \ref{fig13}, when the reprojection error of sight-rays is minimized, the 3D reconstruction error estimated using the corresponding $K$ value is also minimal. This demonstrates that the K value selected by the proposed method achieves the highest 3D reconstruction accuracy. We perform 1000 simulated experiments for each motion pattern and count the accuracy rate of $K$ selection. The experimental results are shown in Table \ref{tab1}. 

\begin{table}[htbp]
    \centering
    \caption{Selection accuracy and calculation speed of our selection algorithm}
    \label{tab1}
    \begin{tabular}{ccccc}
        \toprule
      Ground truth of $K$  & 1 & 2  \\
      \midrule
       Selection accuracy  & 98.1\% & 99.6\%\\
       Average time/s  & $1.57\times 10^{-3}$ & $1.61\times 10^{-3}$ \\
    \bottomrule 
    \end{tabular}
\end{table}

\begin{figure}[htbp] 
\centering
\includegraphics[width=2.5in]{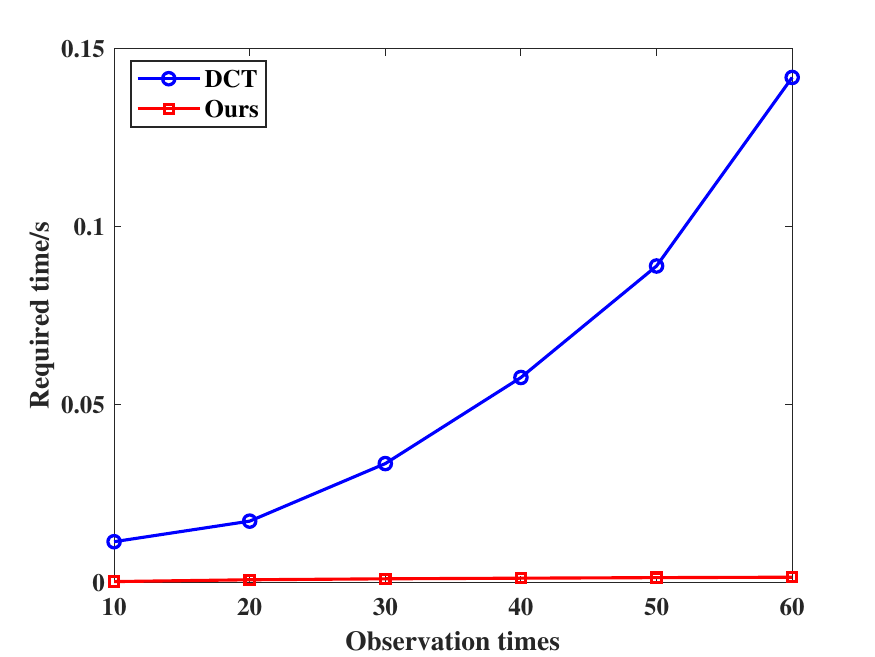}
\caption{Required time of DCT algorithm \cite{Park2015} to select the number of DCT trajectory basis vectors and our algorithm to select the order of temporal polynomials.}
\label{fig14}
\end{figure}

As shown in Table \ref{tab1}, the proposed $K$ selection algorithm in this paper can accurately determine the correct value of $K$. At a high noise level, the selection accuracy is above 98\%. Due to temporal polynomials' significant physical meaning, in a period of time, the order of temporal polynomials for moving vehicles or ships is usually 1 or 2. Therefore, compared with the selection algorithm in reference \cite{Park2015}, our algorithm uses much less time to select $K$. Take the uniform linear motion as an example. As shown in Table \ref{tab1}, the time required for our algorithm to select $K$ is approximately $1.57\times10^{-3}$ s. However, under the same conditions, the DCT algorithm \cite{Park2015} requires $1.42\times10^{-1}$ s to select the number of DCT trajectory basis vectors, which is nearly 90 times that of our algorithm. More seriously, as the number of observations increases, the consumption time of their algorithm exhibits quadratic growth, as shown in Figure \ref{fig14}. This demonstrates the high efficiency and accuracy of the proposed selection algorithm.

\subsection{Performance of handling missing data}\label{sec3.4}

\begin{figure}[htbp] 
\centering
\includegraphics[width=2.5in]{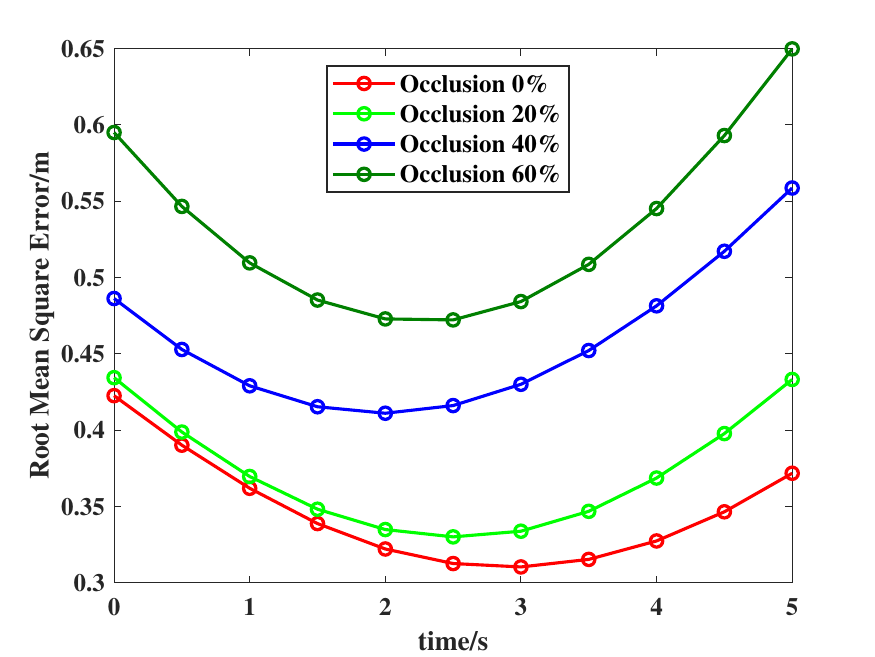}
\caption{Illustration of the experimental result under different occlusion conditions.}
\label{fig15}
\end{figure}

In this section, we evaluate the performance of the proposed method under missing data situations. In real-world scenarios, missing data usually occurs from factors such as motion blur, self-occlusion, or exceeding the field of view. Our algorithm can also handle these missing data situations. To test for the effects of missing data of our algorithm, the simulated experiments are conducted under conditions of high \textit{reconstructability}. We set the position systematic noise of the camera's optical center with a standard deviation of 0.1 m, the position random noise of the camera's optical center with a standard deviation of 0.1 m, the angle systematic noise of the sight-rays with a standard deviation of 0.1°, and the angle random noise of the line of sight with a standard deviation of 0.05°. Occlude 0, 20, 40, and 60\% of the observation data, respectively, and use our algorithm to reconstruct the target trajectory. The mean RMS error under occlusion conditions is shown in Figure \ref{fig15}. The experimental results demonstrate that our method exhibits strong robustness under the weak observation conditions of missing data. It does not degenerate even under up to 60\% occlusion. 
\section{Real-world experiments}\label{sec4}

\begin{figure}[htbp]
\centering
\includegraphics[width=5.5in]{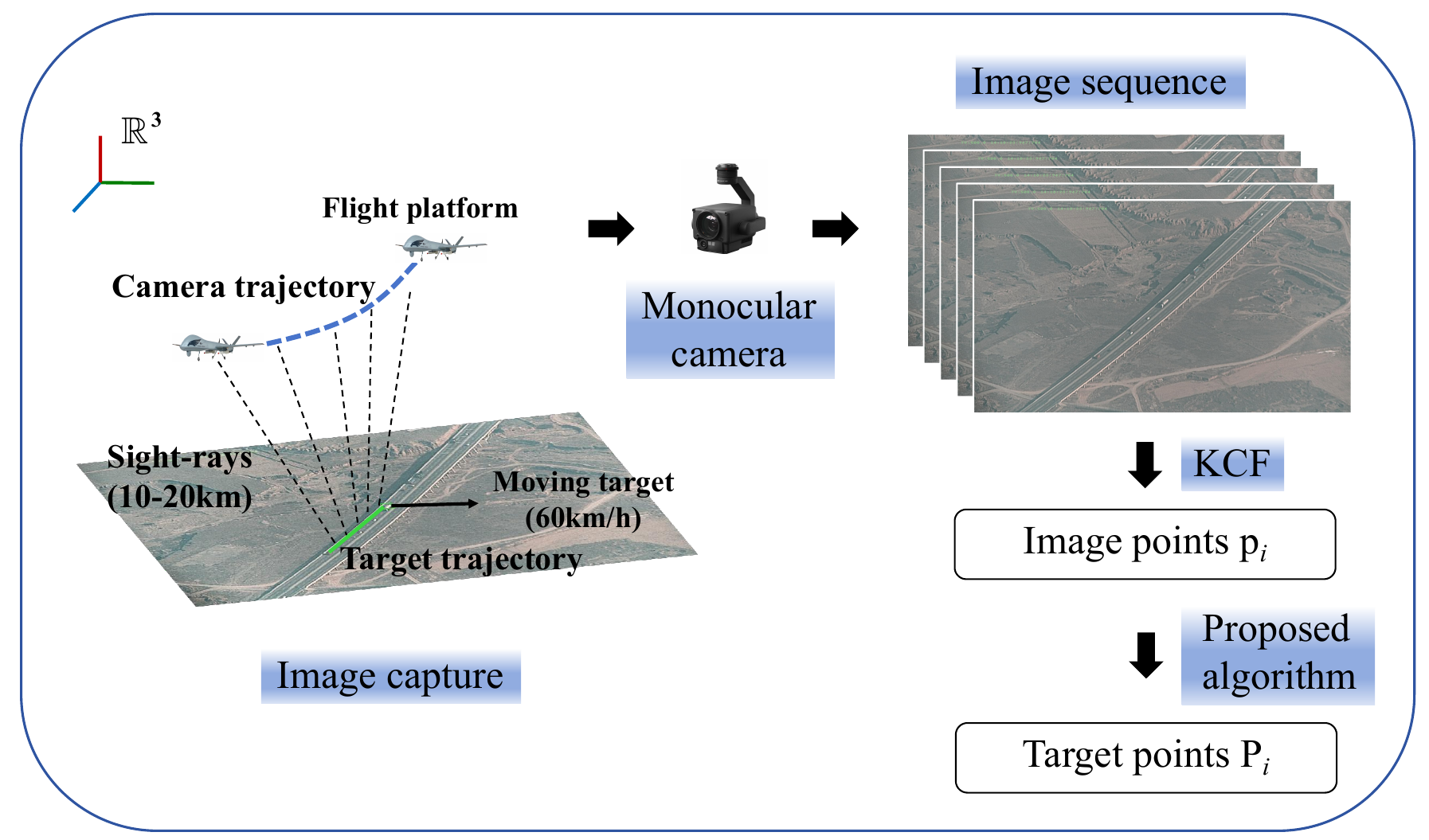}
\caption{Equipped with a monocular camera only, the UAV tracks the moving target and reconstructs its trajectory.}
\label{fig16}
\end{figure}

In this section, we evaluate our method on the real-world data. In the real-world data, the UAV observation platform observes vehicles traveling on the highway. The flight platform is equipped with a monocular RGB camera, which captures images at a spatial resolution of 1280 × 720 pixels and a temporal resolution of 25 Hz. The camera exhibits a constrained field of view, with both its horizontal and vertical angular extents not exceeding 2°. The observation range of the flight platform extends from 10 to 20 km. The moving target travels along the highway in approximately uniform straight-line motion, with a speed of approximately 60 km/h. The position of the camera's optical center and the ground truth of the target trajectory are provided by satellite positioning. Due to the long observation distance, the inclination angle is large. The Kernelized Correlation Filters (KCF) \cite{Henriques2015} algorithm is used to track the target. Under such observation conditions, the target on the road can be regarded as a point target. We are only concerned with the trajectory and motion parameters of the target, regardless of its attitude. The center of the bounding box is taken as the image point of the target. Our purpose is to utilize images captured by aerial platforms equipped with only a monocular camera to reconstruct the target's motion parameters and trajectory, as shown in Figure \ref{fig16}.

\begin{figure}[htbp] 
\centering
\subfloat[]{\includegraphics[width=2.1in]{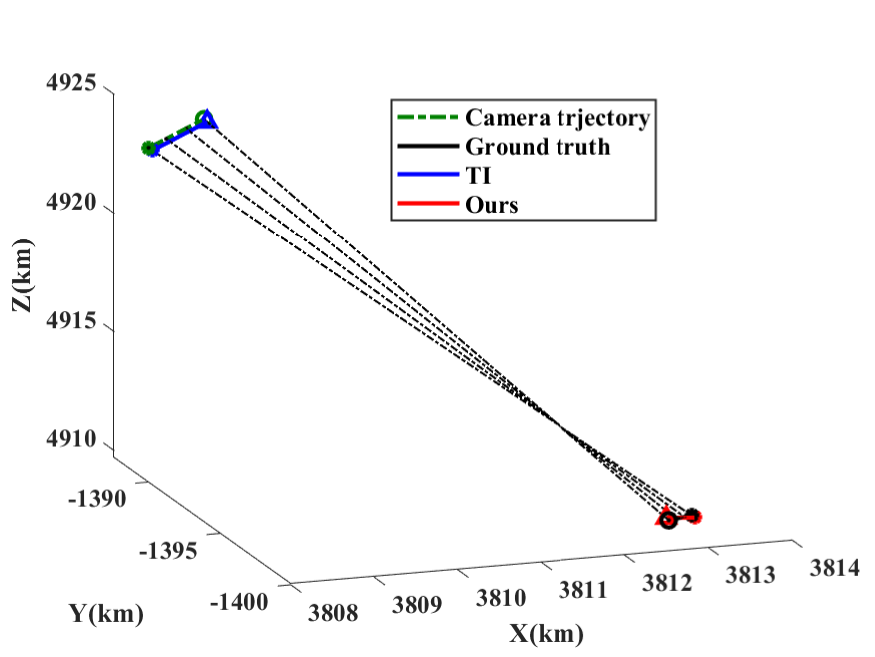}%
\label{fig17a}}
\hfil
\subfloat[]{\includegraphics[width=2.1in]{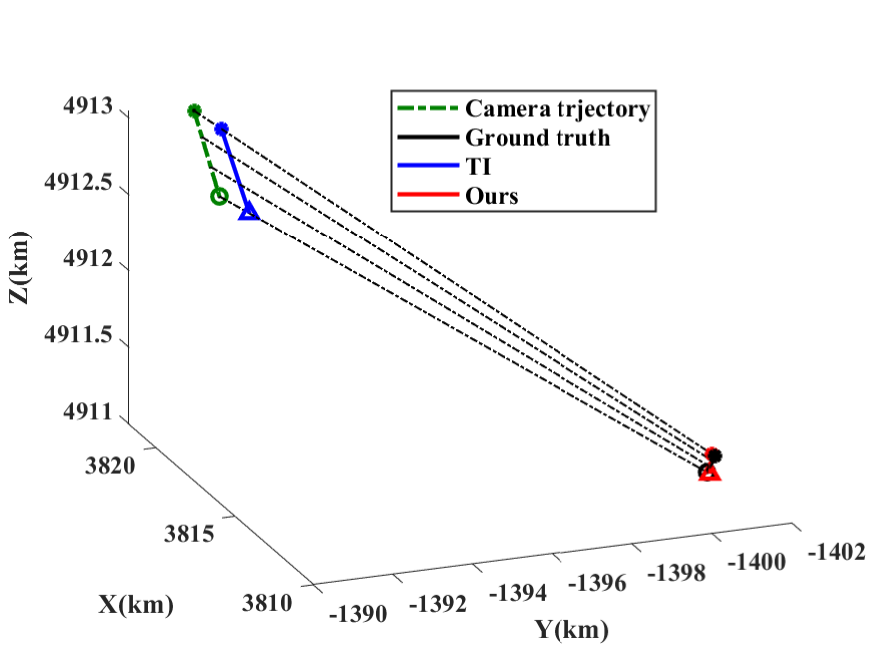}%
\label{fig17b}}
\hfil
\subfloat[]{\includegraphics[width=2.1in]{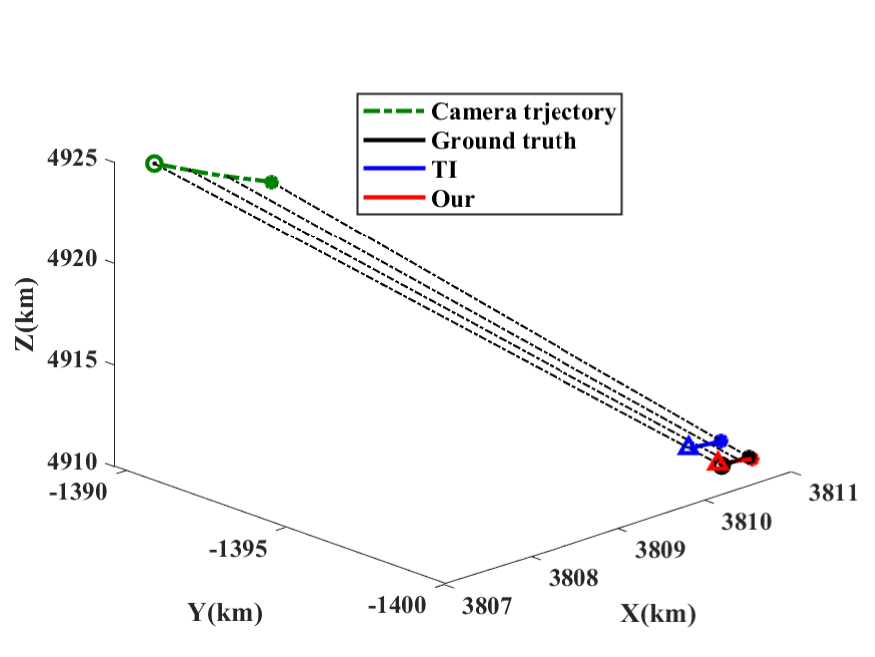}%
\label{fig17c}}
\caption{Illustration of real-world experimental results. (a) Real-world experiment under a low \textit{reconstructability} of 0.29. (b) Real-world experiment under a low \textit{reconstructability} of 0.85. (c) Real-world experiment under a high \textit{reconstructability} of 8.05.}
\label{fig17}
\end{figure}

Three sequences of continuous observational images are selected, each spanning approximately 30 seconds. Therefore, each sequence contains 750 observations, sufficient to solve for the target's uniform linear motion. The flight platform's trajectories are approximated as straight linear over the first two sequences' observation time. Consequently, these conditions approach a degenerate case for the TI method with a low \textit{reconstructability}. In contrast, the flight platform's trajectory is a curve in the last sequence. So, the last observational dataset is conducted under conditions of high \textit{reconstructability}. The values of the \textit{reconstructability} for the three observational datasets are 0.29, 0.85, and 8.05, respectively. The TI method \cite{Yu2009} and the proposed algorithm are employed to reconstruct the target trajectory. Figure \ref{fig17} illustrates the reconstruction results for all three observational datasets. The localization errors are presented in Table \ref{tab2}, \ref{tab3}, and \ref{tab4}, respectively.

\begin{table}[htbp]
    \centering
    \caption{Localization error of Experiment \textbf{a} under the condition of \textit{reconstructability} $\eta=0.29$}
    \label{tab2}
    \begin{tabular}{cccccc}
        \toprule
      Method  & $\mathrm{\sigma_x(m)}$ & $\mathrm{\sigma_y(m)}$ & $\mathrm{\sigma_z(m)}$ & $\mathrm{\sigma(m)}$ \\
      \midrule
      TI  & 4534.25 & 8754.78 & 13741.21 & 16912.32 \\
      Ours  & 15.37 & 30.46 & 49.81 & \textbf{60.37} \\
    \bottomrule 
    \end{tabular}
\end{table}

\begin{table}[htbp] 
    \centering
    \caption{Localization error of Experiment \textbf{b} under the condition of \textit{reconstructability} $\eta=0.85$}
    \label{tab3}
    \begin{tabular}{cccccc}
        \toprule
      Method  & $\mathrm{\sigma_x(m)}$ & $\mathrm{\sigma_y(m)}$ & $\mathrm{\sigma_z(m)}$ & $\mathrm{\sigma(m)}$ \\
      \midrule
      TI  & 10669.07 & 7772.67 & 1241.33 & 13258.37 \\
      Ours  & 37.89 & 27.19 & 4.48 & \textbf{46.85} \\
    \bottomrule 
    \end{tabular}
\end{table}

\begin{table}[htbp] 
    \centering
    \caption{Localization error of Experiment \textbf{c} under the condition of \textit{reconstructability} $\eta=8.05$}
    \label{tab4}
    \begin{tabular}{cccccc}
       \toprule
      Method  & $\mathrm{\sigma_x(m)}$ & $\mathrm{\sigma_y(m)}$ & $\mathrm{\sigma_z(m)}$ & $\mathrm{\sigma(m)}$ \\
      \midrule
      TI  & 148.17 & 540.78 & 714.64 & 908.36 \\
      Ours  & 10.43 & 33.80 & 47.01 & \textbf{58.83} \\
    \bottomrule 
    \end{tabular}
\end{table}

Consistent with the conclusions drawn in Section \ref{sec2.4}, under conditions of low \textit{reconstructability} in the observational dataset, the TI method exhibits degeneracy, with the reconstructed target trajectory closely approximating the camera trajectory. And the smaller the value of \textit{reconstructibility}, the closer the reconstructed trajectory is to the camera trajectory. However, the introduction of ridge estimation effectively mitigates the problem of ill-conditioning, ensuring that our algorithm can still accurately reconstruct the trajectory of the target under conditions of low \textit{reconstructability}. Moreover, under conditions of high \textit{reconstructability} in the observational dataset, the reconstruction accuracy of the TI method remains relatively low due to factors such as long observation distance, narrow field of view, and high noise levels, as shown in Figure \ref{fig17c} and Table \ref{tab4}. In contrast, our algorithm demonstrates superior robustness, enabling the high-precision reconstruction of the target trajectory under such limited observation conditions.

Experiments employing real-world data corroborate the efficacy of our proposed algorithm under limited observation conditions of long observation distance, low \textit{reconstructability}, limited field of view, and high noise levels. The results demonstrate that our method achieves significantly higher accuracy than the conventional trajectory intersection method. Moreover, our method exhibits remarkable stability, maintaining consistent performance without degeneracy. Thus our method shows superior robustness. This experiment also validates the analysis in Section \ref{sec2.4}. As the trajectory of the flight platform becomes more complex, transitioning from straight lines to curves, the \textit{reconstructability} of the system increases, leading to improved reconstruction accuracy.

In practical applications, we recommended maneuvering the UAV in response to the motion of the target points. By adjusting the UAV's steering, acceleration, and deceleration, the complexity of its moving trajectory can be enhanced, thereby enhancing the \textit{reconstructability} of the system. This approach can significantly improve the reconstruction accuracy of the point's trajectory. 
\section{Conclusion}\label{sec5}

This paper proposes an effective method for reconstructing the 3D motion of a point based on a monocular camera. First, temporal polynomials are used to represent the point motion as an additional constraint. Ridge estimation is introduced to the least squares estimation system to mitigate the ill-conditioning caused by limited observation conditions, thereby improving the reconstruction accuracy and robustness. Second, an efficient algorithm is proposed for automatically selecting the optimal order of the temporal polynomials. Third, the geometric relationships among camera motion, target point motion, and the temporal polynomial are analyzed. The \textit{reconstructability} for temporal polynomials is defined to describe the reconstruction accuracy quantitatively. Simulated and real-world experiments demonstrate our method achieves high efficiency, accuracy, and robustness. Especially under limited observation conditions, our method can maintain stable reconstruction results. Future work involves optimizing the UAV's observation trajectory to improve the \textit{reconstructability}.










\bibliographystyle{elsarticle-num}

\bibliography{cas-refs}



\end{document}